\documentclass[final,3p,times]{elsarticle}
\usepackage[colorlinks=true, urlcolor=blue, linkcolor=red]{hyperref}
\usepackage{epstopdf}
\usepackage{epsf}
\usepackage{graphicx}
\usepackage{comment}
\usepackage{mathtools}
\usepackage{amsmath,amssymb}
\usepackage{pdflscape}
\usepackage{amssymb}
\usepackage{amsmath}
\usepackage{bm}
\usepackage{bbm}
\usepackage{subfigure}
\usepackage{booktabs}
\usepackage{multicol}
\usepackage{algorithmic}
\usepackage{algorithm}
\usepackage[dvips]{epsfig}
\usepackage[usenames, dvipsnames, table]{xcolor}
\usepackage{tikz}
\usepackage{array,multirow,makecell}
\usetikzlibrary{calc,shapes,arrows,plotmarks}
\usepackage{layout}
\usepackage{multirow}
\usepackage{color}
\usepackage{afterpage}
\usepackage{balance}
\usepackage[ansinew]{inputenc}
\usepackage{array}
\usepackage{amsmath}

\def\R{\mbox {I\hspace{-.15em}R}}

\colorlet{darkgreen}{black!50!green}
\colorlet{darkblue}{black!50!blue}

\newcounter{LAlgoPerso}

\newcounter{Lcount1}

\newcounter{Lcount2}

\newcounter{Lcount3}

\begin{document}
\begin{frontmatter}
\title{Neuromorphic-based metaheuristics: A new generation of low power, low latency and small footprint optimization algorithms\footnote{This work has been supported by the ERC Generator of the University of Lille.}}
\author{Prof. El-Ghazali Talbi}
\address{Univ. Lille, CNRS, Inria, Centrale Lille, UMR 9189 CRIStAL, F-59000 Lille, France, e-mail: el-ghazali.talbi@univ-lille.fr}

\begin{abstract}
Neuromorphic computing (NC) introduces a novel algorithmic paradigm representing a major shift from traditional digital computing of Von Neumann architectures. NC emulates or simulates the neural dynamics of brains in the form of Spiking Neural Networks (SNNs). Much of the research in NC has concentrated on machine learning applications and neuroscience simulations. This paper investigates the modelling and implementation of optimization algorithms and particularly metaheuristics using the NC paradigm as an alternative to Von Neumann architectures, leading to breakthroughs in solving optimization problems. 

\medskip

Neuromorphic-based metaheuristics (Nheuristics) are supposed to be characterized by low power, low latency and small footprint. Since NC systems are fundamentally different from conventional Von Neumann computers, several challenges are posed to the design and implementation of Nheuristics. A guideline based on a classification and critical analysis is conducted on the different families of metaheuristics and optimization problems they address. We also discuss future directions that need to be addressed to expand both the development and application of Nheuristics.

\end{abstract}

\begin{keyword}
\texttt{Neuromorphic computing, spiking neural networks, neuromorphic optimization, neuromorphic metaheuristics, Nheuristics}
\end{keyword} 
\end{frontmatter}

\section {Introduction}


On one hand, digital computing based on clock-driven architectures faces major barriers due to the end of Dennard scaling \cite{davari1995cmos}, the slowing of Moore’s Law \cite{leiserson2020there}, and the limited bandwidth between CPU and memory—known as the Von Neumann bottleneck or memory wall. These issues result in higher energy consumption, increased latency, and larger hardware footprints. On the other hand, the demand for solving large-scale optimization problems—such as training deep neural networks, large language models, and complex engineering systems—continues to grow, requiring faster and more energy-efficient solutions. As a result, there is an urgent need for novel computing paradigms and optimization methods that minimize power usage, latency, and size. The rapid, often unsustainable growth of AI is also driving up carbon emissions; for instance, training a single large model can emit over 284 tons of CO2 \cite{thompson2021deep}\cite{van2021sustainable}, and AI could account for 0.5\% of global electricity use by 2027 \cite{de2023growing}. Moreover, applications such as edge computing and IoT (Internet of Things), where low latency and small form factors are critical, further underscore the need for more power efficient alternatives \cite{mehonic2022brain}.

\medskip

Neuromorphic Computing (NC) is an emerging and promising alternative to traditional Von Neumann architectures \cite{mead1990neuromorphic}\cite{mead2020neuromorphic}. Inspired by the structure and function of the human brain, NC aims to develop new hardware and algorithms that enable next-generation, non-digital computing paradigms. These systems are designed to provide significant improvements in energy efficiency, reduced latency, and smaller physical footprints (including size and weight). The NC field is inherently interdisciplinary, bringing together researchers from materials science, neuroscience, electrical engineering, and computer science.

\medskip

NC has been proven to be Turing-complete, making it capable of performing general-purpose computing \cite{schuman2021neuromorphic}\cite{date2021computational}. Most of the research work in NC has focused on the development of cognitive applications (e.g., machine learning, computational neuroscience). The application of NC extends well beyond cognitive tasks such as graph algorithms, random walks, partial differential equation (PDE) solving and signal processing \cite{aimone2022review}. These applications highlight the versatility and potential of NC, offering new avenues for solving complex computational problems across various disciplines. 


\medskip

In this paper, we explore the use of NC for the design and implementation of optimization algorithms, with a particular focus on neuromorphic-based metaheuristics (Nheuristics) for tackling challenging optimization problems. NC opens up a new algorithmic landscape for optimization, offering a multi-layered approach that integrates hardware and algorithms to achieve brain-like efficiency. We believe there is a significant algorithmic opportunity for Nheuristics. However, since NC systems are fundamentally different from traditional Von Neumann computers, designing and implementing Nheuristics presents several challenges. We propose a unified framework for defining the key concepts necessary for the design and implementation of Nheuristics. Given the characteristics of Spiking Neural Networks (SNNs) and neuromorphic hardware, new algorithmic directions are required to leverage their implicit recurrence, event-driven nature, and sparse computational features. Novel research is needed to develop neuron models that can process information over time, devise spike-based encodings to represent solutions and data, and design SNN architectures and learning rules suited for sparse, asynchronous, and event-driven dynamical systems.

\medskip

The paper is structured as follows. In section \ref{sec:nc} (resp. \ref{sec:optimization}), the main concepts of NC  (resp. metaheuristics and optimization problems) are provided. Section \ref{sec:design} (resp. \ref{sec:implementation}) defines a general and unified framework for the design (resp. implementation) of Nheuristics. Section \ref{sec:neuro-meta} details a classification and a critical analysis of the different families of Nheuristics (i.e., greedy, local search based, evolutionary and swarm intelligence algorithms). Finally, the last section \ref{sec:conclusion} presents the main conclusions and opens some important research perspectives for the future development of Nheuristics.

\section{Neuromorphic computing}
\label{sec:nc}

In NC, the computational paradigm is inspired from the most efficient information processing system known to humanity: the brain \cite{nilsson2023integration}. NC, which emerged in the late 1980s, focuses on developing sensing and processing systems that mimic the dynamics of the brain \cite{mead1990neuromorphic}\cite{mead2020neuromorphic}. SNNs represent the main algorithmic paradigm for NC that attempts to mimic the synaptic functionalities of the neurons in a temporal, distributed and asynchronous event-driven way. They are considered as the emerging third generation of the neural network model \cite{maass1997networks}\cite{schuman2017survey}\cite{nunes2022spiking}. 

\medskip

Unlike Von Neumann architectures where programs are defined by explicit instructions, programs in NC are defined by the structure of the SNN and its parameters. SNNs use spikes to encode, process and communicate information. The characteristics of those spikes (e.g., time at which they occur, magnitude, shape) will allow to encode numerical information, while in Von Neumann architectures, the information is encoded as numerical values. Another important property of SNNs is their connection plasticity. The synaptic weight between two neurons can be adjusted which enables the SNNs to model and solve complex computational problems. 

\medskip

NC presents some fundamental operational differences with digital computing:
\begin{itemize}
\item {\bf Energy efficiency}: The brain manages billions of neurons and trillions of synapses requiring about 20 watts of power \cite{herculano2011scaling}, while training on a digital computer a large language model (LLM) like GPT-3 requires 1,300 megawatt hours (MWh) of electricity. NC systems are designed to operate on orders of magnitude less power consumption than traditional CPUs or GPUs, making them suitable for battery-operated devices and large-scale implementations. Neuromorphic chips can achieve performance with power consumption as low as milliwatts, compared to watts for conventional systems. Just a single modern GPU such as an NVIDIA Tesla V100, requires about 300 watts. The recent Exascale digital machines are expected to draw tens to hundreds of Megawatts of power.

\item {\bf Massively inherent parallelism}: Neuromorphic computers capitalize on their inherent parallelism, with a large number of neurons and synapses capable of operating in a parallel way. The number of neurons in the
human brain is in the order of 86 billion. Each neuron can establish thousands of synaptic connections with other neurons, amounting to approximately $10^{15}$ synapses across the entire human brain. Compared to parallel Von Neumann systems, their computations and communication are simpler and use reduced precision.

\item {\bf Collocated processing and memory}: Neuromorphic hardware integrates processing and memory functions seamlessly. Neurons and synapses, traditionally viewed as processing units and memory, respectively, perform both computation and data storage in most implementations. This design addresses the Von Neumann bottleneck by eliminating the separation between processor and memory, thereby enhancing throughput. Moreover, it reduces energy consumption by minimizing the need for frequent data accesses to main memory, a significant advantage over conventional digital computing systems.

\item {\bf Scalability}: Neuromorphic computers are built with inherent scalability. By adding more neuromorphic chips, the number of neurons and synapses can be expanded. Multiple physical neuromorphic chips can be integrated to form a larger unified system, enabling the execution of progressively larger SNNs. Neuromorphic architectures can emulate large networks of neurons and synapses efficiently, allowing for more complex computations without a proportional increase in power consumption. This scalability feature has been effectively demonstrated across various prominent neuromorphic hardware platforms like SpiNNaker2 and Loihi2. For instance, the Intel's Hala Point, which is currently the largest neuromorphic system, is composed of 1.15 billion neurons and 128 billions synapses. 

\item {\bf Event-driven asynchronous computation}: Neuromorphic computers utilize event-driven computation, meaning they process data (spikes) only when it is available. This approach leverages the temporally sparse nature of neural activity, where neurons and synapses perform computations only in response to spikes. This strategy ensures highly efficient operation by minimizing unnecessary computational overhead and energy consumption, distinguishing NC systems from traditional clock-based computing architectures. This event-driven nature allows also to respond to stimuli in real time, making them ideal for real-time applications requiring immediate response.

\item {\bf Sparsity}: As noted in \cite{shoham2006silent}\cite{quian2010measuring}, typically fewer than 10\% of neurons in the brain are active at the same time. This contrasts sharply with the inference mode of traditional neural networks, where all neurons are involved in the calculations. This is determined by three factors: temporal sparsity (i.e., data sparse in time), spatial sparsity (i.e., threshold value of the membrane potential) and structural sparsity (i.e., sparseness of the topology of neural connections).

\item {\bf Small footprint}: Neuromorphic chips have the capability to function in environments with strict constraints on size and weight. They can be made compact and lightweight, enabling their use in portable and wearable devices. This miniaturization is particularly beneficial for embedded systems, edge computing and IoT applications.

\item {\bf Stochasticity and chaos}: Experimental data indicates that the brain exhibits inherent stochastic and chaotic behavior \cite{rolls2010noisy}. SNNs can incorporate randomness, for instance, in the firing patterns of neurons, to introduce variability or noise into their operations. 

\item {\bf Local learning}: The training of traditional neural networks relies on the backpropagation algorithm, which needs global operations. In the brain, synaptic weights can be adjusted locally only based on specific activity characteristics of the neurons connected by that synapse.

\item{\bf Robustness to noise and fault tolerance}: The biologically inspired resilience allow to exhibit a level of robustness to noise and can continue functioning effectively in the presence of faults, akin to how biological systems operate. Neuromorphic chips can handle data corruption or inconsistencies better than traditional systems, making them suitable for real-world applications with uncertain conditions. NC architectures often support unsupervised learning, allowing them to adapt to new information and changing environments without the need for extensive retraining.

\end{itemize}

\section{Heuristics and optimization problems}
\label{sec:optimization}

Different families of heuristic algorithms can be considered to be designed and implemented using the NC paradigm. A {\it greedy algorithm} is a heuristic method for solving optimization problems where, at each step, it chooses the best immediate solution without considering future consequences. The idea is to make the best local choice at each step in the hope that these choices lead to an overall optimal solution. However, it is not always guaranteed to find the global optimal solution. Greedy algorithms are commonly used for problems where an approximate solution is acceptable and the computational efficiency is a priority.

\medskip

{\it Metaheuristics} represent a class of general-purpose iterative heuristic algorithms that can be applied to difficult optimization problems \cite{Talbi2009}. One can classifiy metaheuristics into {\it local search} based and {\it population based metaheuristics}. Local search based metaheuristics improve a single solution. They could be seen as search trajectories through neighborhoods. They start from a single solution. They iteratively perform the generation and replacement procedures from the current solution $s$. In the generation phase, a set of candidate solutions are generated from the neighborhood $\mathcal{N}(s)$ of the current solution. In the replacement phase\footnote{Also named transition rule, pivoting rule and selection strategy.}, a selection is performed from the candidate solution set to replace the current solution (i.e., a solution $s^{'} \in \mathcal{N}(s)$) is selected to be the new solution. Popular examples of such metaheuristics are local search (LS) (i.e., hill-climbing, gradient), simulated annealing and tabu search.

\medskip

Population based metaheuristics could be viewed as an iterative improvement of a population of solutions. They start from an initial population of solutions. Then, they iteratively apply some variation operators for the generation of a new population and the replacement of the current population. Using various bio-inspired metaphors, they may be classified into two main categories \cite{Talbi2009}\cite{del2019bio}: 
\begin{itemize}
	\item {{\it Evolutionary algorithms (EAs):} they are guided by the selection and reproduction concepts of survival of the fittest. The solutions composing a population of solutions (i.e., individuals) are selected and reproduced using unary and binary variation operators (e.g., mutation, crossover\footnote{Also called recombination and merge.}). New offsprings (i.e., solutions) are constructed from the different features of solutions belonging to the current population. The most popular EAs are genetic algorithms (GAs), differential evolution (DE) \cite{storn1997DE}, and evolution strategy (ES).}
	
	\item {{\it Swarm intelligence (SI):} they are inspired by the emergence of collective intelligence from populations of agents with simple behavioral patterns for cooperation \cite{eberhart2001swarm}. Particle swarm optimization (PSO) and Ant Colony Optimization (ACO) represent the most popular examples of this class of metaheuristics.}
\end{itemize}

Population based metaheuristics can also be categorized into {\it regular population} and {\it model based population} algorithms \cite{stork2020new}. They are distinct in the way the offsprings are generated during the search. In regular population algorithms, the solutions are generated using population-based variation operators (e.g., mutation and crossover for GAs and DE, and particle update for PSO), whereas in model based algorithms, the solutions are generated by using and adapting a model which stores information during the search such as pheromone matrix in ACO, probability distribution model in Estimation of Distribution Algorithms (EDA), and covariance matrix adaptation in CMA-ES. 

\medskip

The common search components for all metaheuristics which have to be adapted to solve optimization problems are the {\it encoding} of solutions and the {\it variation operators}. Indeed, the encoding (i.e., representation) of solutions is a crucial issue with significant implications on the efficiency of metaheuristics. It defines the decision space in which any metaheuristic will carry out the search process. In general, the initial solutions are randomly and uniformly generated according to the type of variables. The definition of variation operators depends on the family of the used metaheuristic. For local search based metaheuristics, the variation operator is defined by the neighborhood. For regular population based metaheuristics, the variation operators are mutation and crossover (resp. velocity update) for EAs (resp. PSO), while for model-based metaheuristics (e.g., ACO, EDA, CMA-ES), the variation operator is defined by the model building and update.

\medskip

During the last three decades, metaheuristics have been applied with success for a variety of optimization problems. An optimization problem consists to find $x^* = argmin_{x \in X} {f(x)}$, where $X$ is the set of feasible solutions, and $f : X \longrightarrow \R$ is the objective function. In the realm of optimization problems, they can generally be categorized into two main families based on the nature of their variables:
\begin{itemize}
	\item {\it Continuous Optimization Problems (COPs)}: These problems involve variables that can take any real value within a specified range. 
	
	\item {\it Discrete Optimization Problems (DOPs)}: These problems involve variables that can only take discrete values (e.g., integers, permutations). Binary optimization (BOPs) is a special case of DOPs where the variables can only take two values, typically $0$ or $1$.
\end{itemize}

\medskip

The most popular binary (resp. discrete and continuous) problem studied in the neuromorphic community are QUBO (resp. CSP and QP) \cite{glover2022applications}\cite{glover2022quantum}. The general form of a QUBO ({\it Quadratic Unconstrained Binary Optimization}) is $ Min_x x^T Q x $ where $x$ is a binary vector of $n$ variables, $Q$ is $n \times n$ square symmetric or upper-triangular matrix that defines the quadratic terms. QUBO problems are fundamental in various other fields such as operations research, and quantum computing. They can model a wide range of combinatorial optimization problems, including graph coloring, graph partitioning, max-cut, scheduling, and transportation problems. QUBO problems are equivalent to the Ising model in which the binary variables take their value in \{0,1\} instead of \{+1,-1\} \cite{lucas2014}\cite{mohseni2022ising}.

\medskip

A {\it Constraint Satisfaction Problem} (CSP) is a type of DOP where the goal is to find an assignment of values to a set of variables that satisfies a set of constraints. CSPs are typically used in areas like artificial intelligence, scheduling, and resource allocation. A CSP is defined by the following components: A set of variables \( V = \{v_1, v_2, \dots, v_n\} \) ;  a domain \( D(v_i) \) for each variable \( v_i \), which specifies the possible values that \( v_i \) can take ; and a set of constraints \( C = \{c_1, c_2, \dots, c_m\} \), where each constraint \( c_i \) restricts the values that a subset of variables can take simultaneously. The goal is to find an assignment of values to the variables such that all constraints in \( C \) are satisfied.

\medskip
A {\it Quadratic Programming} (QP) problem is a type of COP where the objective function is quadratic, and the constraints are linear. Formally, it is defined as: minimize $ f(x) = \frac{1}{2} x^T Q x + p^T x $, subject to $A x \leq k$, where $A \in \R^{M \times L}$, $x \in \R^L$, $p \in \R^L$, $k \in \R^M$ and $Q \in \R^{L \times L}$. This also includes the special case of linear programming (LP), where $Q=0$. The problem can be solved using first-order gradient descent $ x_{t+1} = x_t - \alpha . \triangledown f(x) = (1-\alpha Q ) x_t - \alpha p $, where $\alpha$ defines the step size of the gradient descent.

\section{Design of Nheuristics}
\label{sec:design}

Experimental \cite{berkes2011spontaneous} and theoretical \cite{buesing2011neural} studies propose modeling SNNs as Markov chains rather than Turing machines. Markov chains function as finite state machines that transition between states based on probabilistic rules. For instance, if a network of $N$ spiking neurons is represented as a Markov chain, its state at time $t$ can be described as a binary vector of size $N$, where each entry is $1$ if the corresponding neuron fires within a small time window of length $\tau$ (e.g., $\tau = 10ms$). The stationary distribution $p$ of the Markov chain can be interpreted as the analog ``program'' guiding the SNN to perform a specific type of computation. Hence, Markov chains can be utilized to generate Nheuristics for solving optimization problems. 

\medskip

The benefits of employing the neuromorphic paradigm in the design of efficient Nheuristics can be summarized as follows:
\begin{itemize}
\item Local data access supports fast iterative updates of neuronal states (i.e., solutions, populations). Each neuron (e.g., decision variable) computes on its own memory.
\item Parallel computation of the objective function.
\item Parallel generation of solutions and populations, by updating the neuronal states for all variables simultaneously.
\item Parallel selection and replacement of solutions.
\end{itemize}

\medskip

The main concepts of SNNs needed to be defined for the design of Nheuristics are: neuron model, information encoding, SNN architectures and learning rules.

\subsection{Neuron model}

\begin{figure}[h]
\centering
\includegraphics[scale=0.5]{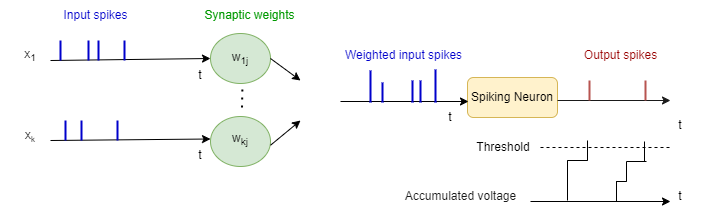}
\caption{The principle of a spiking neuron in SNNs.}
\label{fig:spiking-neuron}
\end{figure}

Neurons are the core building blocks of the brain. Each neuron is made up of a cell body, dendrites, and an axon. Dendrites receive signals from other neurons and transmit them to the cell body for processing. The axon carries signals away from the neuron to communicate with other neurons. Neurons transmit information via electrochemical signals. Synapses are the junctions where one neuron's axon connects to another neuron's dendrite. When a neuron receives signals through synapses, an electrical charge builds up across its membrane, changing its voltage potential. Once the voltage reaches a certain threshold, the neuron fires an action potential, an electrical impulse that travels down the axon and influences other neurons at their synapses. A neuron's behavior is governed by processes such as charge accumulation and the firing mechanism, which involve the dynamics of the cell membrane, ion channels, axons, and dendrites. Figure \ref{fig:spiking-neuron} illustrates the dynamics of a spiking neuron, with input spikes on the left, membrane dynamics in the center, and spiking outputs on the right.

\medskip

The definition of the neuron model is a key question in SNNs \cite{Izhikevich}. Most spiking neuron models share several common features \cite{shrestha2022survey}. They maintain an internal state that accumulates input stimuli, fire or generate an output when this internal state surpasses a threshold value, and often include a {\it refractory period} following firing during which the neuron remains dormant and does not respond to further input stimuli. Various spiking neuron models have been proposed, each balancing between {\it biological accuracy} and {\it computational feasibility}. They can be classified according to their biological plausibility (i.e., mimicking neuroscience) (Fig.\ref{fig:neuron-models}):

\begin{figure}[h]
\centering
\includegraphics[width=0.8 \linewidth]{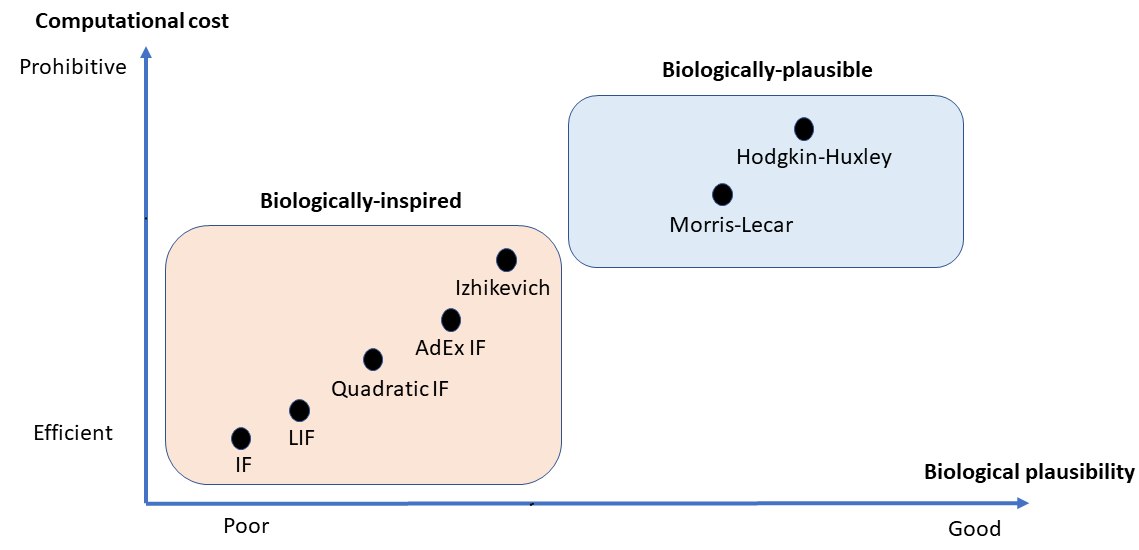}
\caption{Comparison of neuron models in terms of biological plausibility and computational complexity.}
\label{fig:neuron-models}
\end{figure}

\begin{itemize}	
\item {\it Biologically-inspired neuron models}: Those models try to replicate the behavior of biological neural systems but not in a biologically-plausible way. The most simple biologically-inspired model is the integrate-and-fire (IF) model \cite{abbott1999lapicque}. It is defined by the following differential equation:
$$
\begin{array}{l l}
	C\frac{\partial{V}}{\partial{t}}=I & \text{If }  V < V_{th} \\
	V=V_{reset} & Else 
\end{array}
$$
where $C$ represents the membrane capacitance, $I$ the input current (spikes), $V_{th}$ the threshold and $V$ the membrane potential. Once $V$ reaches $V_{th}$, the $IF$ neuron fires a spike at its output. The most popular model is the Leaky integrate-and-fire (LIF) model \cite{gerstner_kistler_2002}, in which the main operations are: synaptic integration $V_j(t)=V_j(t-1)+ \sum_{i=1}^{N-1}{x_i(t)s_i}$, leak integration $V_j(t)=V_j(t)- \lambda_j$, threshold, spike firing, and reset: if $V_j(t) \geq \alpha_j$ spike and $V_j(t)=R_j$, where $V_j(t)$ represents the summation of the membrane potential of the neuron $j$ at time $t$, $x_i(t)$, $s_i$ the synaptic inputs and weights, $\lambda_j$ the {\it leak value} which is substracted from the neuron's potential at each time-step, $\alpha_j$ the threshold voltage and $R_j$ the reset value of the voltage. 


\medskip 

The next level of complexity beyond basic IF models includes the general nonlinear IF models, such as the quadratic IF model employed in some NC systems. Another step up in complexity is seen with the adaptive exponential (AdEx) IF model, which approaches the level of complexity found in models like the Izhikevich model. The Izhikevich model presents the firing behavior of a neuron with an arithmetic complexity close to the LIF model \cite{Izhikevich}. The Izhikevich model has gained popularity in neuromorphic literature due to its simplicity and capacity to replicate biologically accurate neuronal behavior simultaneously \cite{schuman2017survey}. It provides a balanced trade-off between biological realism and computational efficiency. The quadratic and the Adaptive Exponential (AdEx) IF models are a transformation of the Izhikevich model, in which it reduces the response speed of the neuron's membrane voltage \cite{brette2005adaptive}.

\item {\it Biologically-plausible neuron models}: those neuron models explicitely use the behavior of biological neural systems. They include other biologically-inspired components such as dendrites, axons and glial cells. The most widely recognized biologically plausible neuron model is the Hodgkin-Huxley (HH) model \cite{burkitt2006review}. The HH model is a relatively complex neuron model and computationally expensive that describes neuron behavior using four-dimensional non linear differential equations. A simpler yet biologically plausible model is the Morris-Lecar model, which reduces neuron dynamics to a two-dimensional nonlinear equation \cite{lecar2007morris}.
\end{itemize}


\medskip

The deterministic neuron model emits a spike immediately when the membrane potential crosses the threshold voltage. A probabilistic neuron model fires in a stochastic or chaotic way, meaning the probability of firing at any given time is a non-linear function of the instantaneous magnitude of the weighted input. This stochastic/chaotic firing behavior introduces variability in the neuron's response to inputs.

\medskip

{\bf Neuron models and Nheuristics}: Biologically plausible models are computationally expensive and often not effective for Nheuristics. Today, simplified biologically-inspired models, such as Leaky Integrate-and-Fire (LIF) or Integrate-and-Fire (IF) models, are commonly used to achieve more efficient designs. As a result, much of the existing work on Nheuristics relies on LIF models due to their simplicity in computation and lower power consumption \cite{alom2017quadratic}. Furthermore, models like LIF and IF facilitate accurate and efficient hardware implementations. Moreover, hardware-based implementations often limit the types of neuron models that can be used. For example, hardware systems like TrueNorth and Loihi are designed to implement only the LIF model. Another key consideration is the use of random or stochastic models. Since many metaheuristics are stochastic algorithms, various stochastic \cite{kang2024solving} and oscillator-based neuron models \cite{fang2019swarm} have been explored in the literature. However, to the best of our knowledge, chaotic models, which are more biologically plausible \cite{mohseni2022ising}, have not been yet investigated in the context of designing Nheuristics.

\subsection{Information encoding}

SNNs enable highly efficient computations by leveraging their ability to encode information both spatially and temporally. Encoding information into spikes in a way that mimics biological processes remains one of the most significant challenges in neuroscience. Neural encoding continues to be a key area of research for both neuroscientists and computational AI researchers. A spike represents a binary signal, and data such as real numbers or images must be encoded into a series of spikes distributed over a time window. Various methods exist for encoding information using spatio-temporal combinations of uniform spikes. Therefore, developing effective ways to encode complex information into spikes is essential for advancing SNNs, as these networks rely on spikes and their sequences to convey information \cite{zenke2021visualizing}\cite{auge2021survey}. In the literature, two primary approaches for information encoding are identified: {\it rate encoding} \cite{Butts2007} and {\it temporal encoding} \cite{panzeri2010sensory}\cite{guo2021neural}.

\subsubsection{Rate encoding}

\begin{figure}
\centering
\includegraphics[scale=0.55]{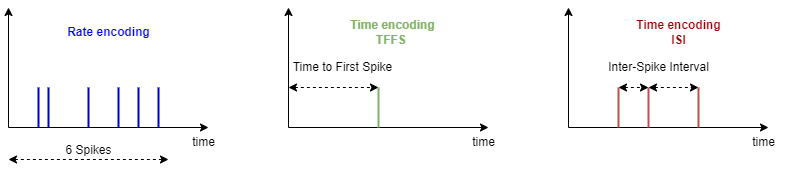}
\caption{Temporal diagram showing the number of emitted spikes based on the type of encoding used in SNNs: rate encoding, time encodings TTFS and ISI.}
\label{fig:encodings}
\end{figure}

The rate encoding scheme is based on the frequency of spikes produced over a given time window. The timing of the spikes has no relevance. Information is encoded by counting the spikes generated within that time window (Fig.\ref{fig:encodings}). The faster a neuron fires, the higher the value it represents, and vice versa. A longer window allows for more accurate rate estimation but results in slower processing, whereas a shorter window can make the network more responsive but less precise. The most popular rate encoding scheme is the count encoding in which the count rate is defined by the mean firing rate $ v=\frac{N_{spike}}{T} $, where $N_{spike}$ is the count of spikes and $T$ a time window. 


\medskip

Rate coding is a straightforward and effective method for encoding information in SNNs, making it particularly popular in scenarios where latency is not a primary concern. It is simpler to implement and analyze compared to other encoding schemes, such as temporal coding. Additionally, rate coding is more robust to noise in spike timing; as long as the overall firing rate remains constant, the information is preserved, even if the exact timing of individual spikes fluctuates.

\medskip

However, rate encoding has several limitations, including latency, inefficiency, and potential information loss, especially when compared to temporal encoding methods. Accurate rate estimation requires multiple spikes, which can introduce delays in processing and reduce the network's speed. Additionally, rate encoding can be energy-inefficient, as it requires high spike rates to represent large values, leading to higher energy consumption and greater computational demands. Furthermore, rate encoding ignores the precise timing of individual spikes, which can be important in certain cases, while temporal encoding can convey more information using fewer spikes.

\subsubsection{Temporal encoding}

Some neuroscientists suggest that a portion of information in the brain is encoded through precise timing of spikes \cite{thorpe1996speed}. The temporal encoding scheme focuses on the exact timing of spikes, with earlier spike times representing more significant information. Unlike rate encoding, temporal encoding generates much sparser spikes, as the timing of the spikes—rather than their frequency—carries the information. This approach allows SNNs to leverage the dynamic nature of spike sequences, enabling faster and more efficient computations. Temporal coding operates on a fine timescale, typically in the millisecond range, which allows for encoding more information in shorter time intervals, facilitating rapid processing and efficient use of spikes.

\medskip

Temporal encoding can be implemented in several forms \cite{auge2021survey}. The most popular ones are (Fig.\ref{fig:encodings}):
\begin{itemize}
\item {\bf Time To First Spike (TTFS)}: In TTFS, we encode the analog value based on the time it takes for the first spike to arrive \cite{rueckauer2018conversion}. 

\item {\bf Inter-Spike Interval (ISI)}: In ISI, we encode information by measuring the interval between spikes. It refers to the precise delay between consecutive spikes to encode the intensity of activation \cite{kumar2010spiking}. 
\end{itemize}

\medskip

Temporal encoding enables high information density with fewer spikes. By utilizing precise timing to encode information, it requires fewer spikes than rate coding, making the system more energy-efficient. Temporal coding can also operate much faster than rate coding, as information can be conveyed in a single spike or a few spikes at specific times, eliminating the need to wait for multiple spikes to accumulate for rate computation. This approach supports more complex and dynamic patterns of neural activity, making it ideal for encoding rapidly changing information, such as in auditory processing or real-time sensory systems.
\medskip

Temporal coding is more complex to implement and analyze than rate coding. The need for precise timing requires careful control of spikes, making the development and training of temporal-coded SNNs more challenging. Temporal coding is also more sensitive to noise in spike timing—if the timing of spikes is disrupted by noise, the encoded information can be lost or corrupted, while rate coding is generally more robust to such fluctuations. Achieving high precision in both time resolution and synchronization is essential for effective temporal coding, which can place additional demands on neuromorphic hardware and algorithms.




\medskip

{\bf Information encodings and Nheuristics}: The optimal encoding strategy is still an open question, necessitating further exploration of how it interacts with neuron models and SNN architectures. The choice of an encoding scheme is likely to be task-specific and subject to optimization \cite{guo2021neural}\cite{forno2022spike}\cite{schuman2022evaluating}. Binary code, where information is represented by which neurons fire during a temporal interval, is well adapted to solve BOPs. Consequently, Nheuristics are particularly well-suited for solving BOPs. However, it is possible to encode discrete and continuous values as spikes using spatio-temporal encoding schemes \cite{kiselev2016rate}\cite{pan2019neural}\cite{schuman2019non}.

\medskip

Just like in biological systems, certain encoding techniques are more effective than others, depending on the type of data and the SNN architecture involved. Rate encoding remains a dominant choice in Nheuristics due to its straightforward implementation and robust performance. It would require more than a single spike per neuron—and thus a longer duration and more power — to represent more information. Without increased latency or a higher spiking frequency, rate coding is prone to significant quantization errors. Additionally, when spikes are generated as stochastic events, sampling errors may also arise. It will be essential to incorporate some level of time coding in Nheuristics to tackle large scale optimization problems. Taking spike timing into account significantly enhances the information capacity of a spike train. Therefore, the selection of a encoding scheme will likely be algorithm and problem-specific and subject to optimization.

\medskip

When data isn't directly encoded as spike trains, it must be converted into spikes. The most common method in rate coding is Poisson coding, which translates a digital value into a spike frequency—higher digital values result in higher frequencies \cite{heeger2000poisson}. However, a significant drawback of this approach is that it requires a large amount of spiking activity to achieve good predictive performance, and an increase in spikes can lead to reduced latency \cite{oh2021spiking}. The most widely used temporal conversion method is TTFS. Within a designated encoding time window, TTFS encodes higher digital values as earlier spikes, while lower values are represented by later spikes. Another technique, known as rank-order coding, conveys information based on the order of the spikes rather than their exact timing \cite{thorpe1998rank}. In temporal ISI encoding, the temporal distance between two spikes can represent a numeric value.  

\medskip

Neuron firing synchrony has been observed in the visual cortex and is associated with dynamic interactions within the network \cite{singer1999neuronal}. This suggests that information is not only encoded through individual neuron firing rates or timing but is also communicated at a population level through the simultaneous activity of multiple units. Some researchers suggest that SNNs could leverage this insight to enhance their computational capabilities, particularly for datasets where the timing of inputs is crucial to the application. In {\it Population encoding} the information is represented collectively by a group of neurons rather than a single neuron. This distributed representation enhances robustness, increased precision and computational efficiency. 

\medskip

Rate and time encodings are both considered biologically plausible. In the brain, neurons may use a mix of both rate and temporal encodings to transmit information \cite{auge2021survey} and the brain employs various encoding strategies tailored to different tasks \cite{Zador1998ImpactOS}\cite{eliasmith2003neural}. Hybrid approaches, such as burst coding, combine spike frequency with inter-spike intervals. Research has shown that burst coding improves performance in SNNs for visual tasks \cite{park2019fast} and enhances energy efficiency in neuromorphic hardware \cite{chen2017real}.

\subsection{SNN architectures} 

\begin{figure}
	\centering
	\includegraphics[scale=0.5]{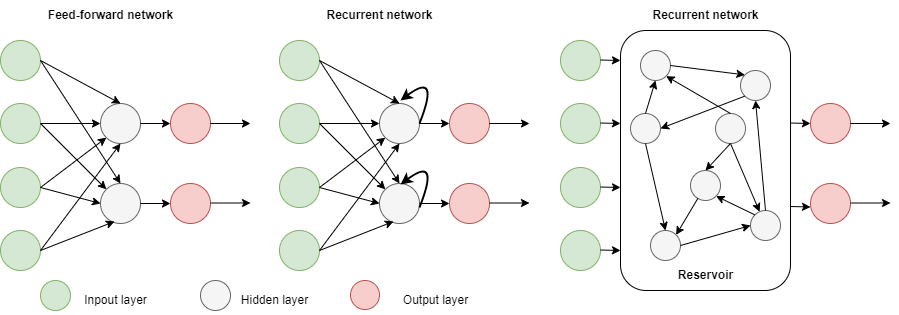}
	\caption{Feedforward and recurrent SNN architectures.}
	\label{fig:topologies}
\end{figure}

SNN architectures detail the connections and interactions between different neurons and synapses. One can consider two main families of architectures: feed forward and recurrent (Fig.~\ref{fig:topologies}). By far, the most popular SNN architecture is feed-forward SNNs (FFSNNs). Information flows unidirectionally from the input layer, through hidden layers, to the output layer, similar to a traditional feedforward neural network (e.g., MLPs). Recurrent SNNs (RSNNs) incorporate feedback connections, where neurons in one layer can influence neurons in the same or earlier layers. RSNNs  can store temporal information, making them powerful for time series and sequence-based tasks. Liquid State Machines (LSMs) is a type of RSNN where a reservoir (i.e., a large recurrent network) processes inputs into a high-dimensional dynamic state, from which a simple readout layer extracts features for tasks like classification or regression. LSMs or reservoir networks are a specific type of SNNs inspired by the brain. Unlike FFSNNs organized in distinct layers, LSMs have a different topology (Fig.\ref{fig:topologies}). They consist of neurons that are randomly interconnected, forming a reservoir where each unit can receive input spikes from both external inputs and other neurons within the reservoir. This random and recurrent connectivity allows LSMs to exhibit complex dynamic behaviors and perform tasks such as temporal processing. Due to their complex computational graph representation, reservoir networks may not efficiently map onto current neuromorphic hardware.

\medskip

\medskip~

{\bf SNN architectures and Nheuristics}: When choosing an SNN architecture, there are many options to consider, depending on the design of the Nheuristic and the target optimization problem. While conventional feedforward SNNs have been mainly used for greedy Nheuristics, iterative Nheuristics (e.g., local search, EAs, SIs) incorporate recurrence in the SNN, where the outputs of the SNN serve as inputs for the next iteration. Nheuristics have mainly been applied to small and medium-sized optimization problems, with complex architectures like deep, reservoir, and hierarchical models remaining largely unexplored.

\medskip

In general, handcrafted SNN architectures are designed to tackle specific optimization problems. For instance, when solving QUBO and graph-related problems, the input data (i.e., the graph) can be directly mapped onto the SNN architecture, with synaptic delays determined by edge weights \cite{kay2020neuromorphic}. However, only a few Nheuristics are built on generic SNN architectures capable of handling a diverse range of optimization problems.

\subsection{Learning rules} 


Compared to ANNs, SNNs are more difficult to train because of their complex dynamics and the non-differentiable nature of spikes. The exact learning rule in which a neuron is trained in the brain is an open question. Learning rules for SNNs can be classified into:

\begin{itemize}
\item {\bf Unsupervised Learning}: Neuroscientists have identified the spike-time-dependent plasticity (STDP) rule for learning. STDP is a widely used local learning rule where the timing difference between pre-synaptic and post-synaptic spikes determines synaptic weight changes \cite{bi1999distributed}. Neurons that spike together frequently strengthen their connections, similar to Hebbian learning ("cells that fire together wire together"). Hebbian Learning reinforces connections when there is simultaneous activity between neurons. This is especially useful for discovering patterns in unsupervised learning settings.

\item {\bf Supervised Learning}: Because spikes are non-differentiable, gradient-based methods (e.g., backpropagation) cannot be directly applied to SNNs. Surrogate gradients approximate the gradient of the spike function to train deeper SNNs. Another training method is based on ANN-to-SNN conversion. This method converts pre-trained ANN architectures into SNNs by mapping continuous neuron activations to spiking rates, allowing for deep learning models to operate as spike-based networks.

\item {\bf Reinforcement learning}: SNNs can incorporate reinforcement learning through a combination of STDP and reward signals. This allows the network to adjust synaptic weights based on the timing of spikes and external reward feedback.
\end{itemize}

{\bf Learning rules and Nheuristics}: Most of the proposed Nheuristics use unsupervised learning rules. STDP and stochastic noise are generally carried out to generate new solutions. Inspired by the brain, we can also distiguish two types of neurons in a SNN:
\begin{itemize}
\item {\it Excitatory neurons} generate a positive action potential that stimulates other connected neurons.
	
\item {\it Inhibitory neurons} generate a negative action potential, which reduces the membrane potential of other connected neurons. When inhibition is strong enough to prevent other neurons from spiking, it can implement a Winner-Takes-All (WTA) mechanism, ensuring that only one neuron spikes at a time \cite{maass2000computational}. In this process, units within a single layer compete, and the unit receiving the strongest input spikes and inhibits all others. The WTA mechanism, inspired by brain function, plays an important role in Nheuristics \cite{amari2013competition}\cite{salzman1994neural}. A variant of the WTA mechanism is the $k$-WTA, which allows up to $k$ neurons in a layer to fire before full inhibition occurs.
\end{itemize}



\subsection{Computational complexity of Nheuristics}

A big challenge in neuromorphic algorithms is the establishment of a theoretical framework to define their computational complexity \cite{kwisthout2020computational}\cite{date2021computational}. One has to define the space and time complexity of Nheuristics. The {\it space complexity} of a Nheuristic can be defined by the number of neurons and synapses used in the associated SNN. The number od synapses of a SNN is bounded by $O(N^2)$ where $N$ is the number of synapses. In this worst case scenario, all neurons are connected to each other. The {\it time complexity} of Nheuristics consists of two components: setup time and run time. Setup time refers to the cost of establishing the SNN, which involves configuring all the neurons and synapses that comprise it. Consequently, it is proportional to the space complexity of the algorithm. Run time is the duration required to execute the SNN, starting from when the inputs are received to when the outputs are generated. In the absence of synaptic delays, the worst-case run time can be assessed by determining the number of neurons and synapses along the longest path from an input neuron to an output neuron. Since the number of synapses will equal the number of neurons in this longest path, calculating either metric suffices for estimating run time. If synaptic delays are present in the SNN, they must also be factored into the run time calculation. Once both setup time and run time are determined, the overall time complexity can be calculated by summing them in the standard $big-O$ notation.

\section{Implementation of Nheuristics}
\label{sec:implementation}

Nheuristics can be implemented on a given neuromorphic hardware or simulated using various software tools.

\subsection{Neuromorphic hardware}

Neuromorphic hardware refers to computational architectures and devices engineered to emulate the structure and function of the human brain (see Tab.~\ref{tab:architectures}). In hardware design, achieving an optimal balance among cost, flexibility, performance, and energy efficiency is often challenging due to trade-offs between these objectives. These architectures are typically classified into three main categories based on their implementation: digital, analog, and mixed-mode systems \cite{shrestha2022survey}.

\subsubsection{Digital neuromorphic architecture}
                                            ~
Digital neuromorphic hardware uses traditional binary logic (0s and 1s) to simulate the behavior of neurons and synapses. They often rely on event-driven architectures where computations are triggered by the occurrence of spikes, rather than a global clock. The key features of digital hardware are their event-driven (asynchronous) processing, digital logic (transistors, gates), simulations of neuron and synapse behavior, and support of large-scale SNNs. It can further be classified as:

\begin{itemize}
\item {\bf CPU based}: These systems utilize traditional central processing units (CPUs) for their computational tasks. An example is SpiNNaker which is composed of thousands of ARM cores and a specific interconnect network optimized for SNNs \cite{furber2020spinnaker}\cite{painkras2013spinnaker}. SpiNNaker2 enables the modeling of 152,000 neurons and 152 million synapses with 152 ARM cores, consuming between 2 and 5 watts. 

\item {\bf ASIC based}: Application Specific Integrated Circuit (ASIC) implementations are custom-designed chips tailored specifically for NC. Examples include IBM's TrueNorth \cite{debole2019truenorth}\cite{akopyan2015truenorth}, Intel's Loihi \cite{davies2018loihi}, ODIN \cite{frenkel20180} and Tianjic \cite{pei2019towards}. They are subject to limitations of specific neuron models. The Loihi2, which operates approximatively at 1 watt, consists of 6 embedded microprocessor cores (Lakemont x86) and 128 fully asynchronous neuron cores (NCs) connected by a network-on-chip. A single Loihi2 chip supports up to 1 million neurons and 120 million synapses \cite{orchard2021efficient}. Several neurocomputers with varying capabilities have been developed based on Loihi, with Pohoiki Springs being the most powerful. This system comprises 768 Loihi chips organized into 24 modules on a single motherboard, effectively simulating 100 million neurons.
	
\item {\bf FPGA based}: Field-Programmable Gate Arrays (FPGAs) are reconfigurable hardware platforms used in some digital neuromorphic systems. FPGA has drawn much attention, as they offer flexibility in design and can be optimized for specific SNN models \cite{javanshir2022advancements}.
\end{itemize}

\subsubsection{Analog neuromorphic architecture}

These systems use analog circuits to closely replicate the behavior of neurons and synapses, mimicking biological systems. Analog hardware has the potential to provide advantages in power efficiency and computational speed for specific neural network tasks. Analog NC systems rely on continuous electrical signals (voltages and currents) to directly emulate the dynamics of biological neurons and synapses, leveraging the inherent physics of electronics to approximate biological processes.

\medskip

The key features of analog hardware include continuous-time, analog computation, direct emulation of neurons and synapses, and low power consumption. Analog hardware is particularly well-suited for simulating complex neural behaviors, such as graded potentials and ion channel dynamics. However, it can be more challenging to scale and program, and its precision may be limited by noise and variability in components.

\subsubsection{Hybrid analog/digital neuromorphic architecture}

Hybrid-mode platforms integrate analog components (e.g., for neuron and synapse dynamics) with digital components (e.g., for programmability and network management). The goal is to leverage the advantages of both circuit types, using analog circuits for efficient neural emulation and digital circuits for computation and control. These platforms combine the realism, speed, and low power consumption of biological neural networks through analog circuits with the flexibility and programmability provided by digital components. Notable examples of such NC systems include BrainScaleS \cite{grubl2020verification}, BrainDrop, NeuroGrid \cite{benjamin2014neurogrid}, and DYNAP-SEL \cite{schuman2017survey}. The main challenge in hybrid-mode hardware is balancing the analog and digital components, ensuring proper synchronization, and minimizing noise in the analog sections.

\subsubsection{Emerging neuromorphic devices}

At the device level, the emerging technologies in nano devices and materials represent cutting-edge technologies designed to mimic the brain's architecture and processing capabilities. They provide a potential for extremely small, ultra-fast and extremely low-lower neuromorphic hardware if they are successfully married with suitable algorithms. Some notable examples include:

\begin{itemize}	
\item {\bf Memristive devices (Non-volatile Memory)}: Emerging nanoscale devices, such as resistive memory or memristive devices, offer non-volatile storage capabilities that are especially suitable for constructing computing substrates for SNNs \cite{panda2024recent}. These devices store information as a physical state in their resistance or conductance states. The four primary types of memristive devices include Phase Change Memory (PCM), Metal Oxide-based Resistive Random Access Memory (RRAM), Conductive Bridge RAM (CBRAM), and Spin-Transfer-Torque Magnetic RAM (STT-MRAM) \cite{mu2020organic}. Memristors have an analog-like behavior for SNNs and naturally implement synaptic plasticity by adjusting their resistance based on the history of current and voltage, emulating learning processes. Currently, neuromorphic memristor computing systems have not yet reached a production-ready state. However, it's worth highlighting several significant research projects that showcase the distinctive neuromorphic properties of memristor approaches \cite{yao2020fully}\cite{li2018efficient}.

\item {\bf Optical devices (photonic)}: Optical neuromorphic systems use photonic circuits for high speedup signal transmission and processing, rather than electrons to process information, potentially offering faster and more energy-efficient processing than conventional electronic systems. They can encode spikes through light pulses. Moreover, they have an extremely low energy consumption and high bandwidth. 
 
\item {\bf Ferroelectric devices}: Ferroelectric neuromorphic devices leverages the unique properties of ferroelectric materials to emulate neural functions, such as memory and synaptic behavior. They are characterized by their non-volatile polarization states, low power consumption, and high endurance.
	
\item {\bf Optoelectronic devices}: Optoelectronic devices integrate optical and electronic components to mimic SNNs. They can handle both light and electrical signals, providing enhanced flexibility and speed.
	
\item {\bf Magnetic devices}: Magnetic devices utilize magnetic properties to store and process information, resembling the functions of synapses and neurons. They enable high-speed data processing while maintaining low energy consumption.
\end{itemize}

\medskip

\begin{table}
\begin{tabular}{| m{3cm} | m{4cm} | m{4cm} | m{4cm} |}
\hline
Category &	Key Features &	Advantages &	Examples \\
\hline
Digital architecture &	Digital spiking neuron models, event-driven	& Programmable, scalable, large networks &	IBM TrueNorth, Intel Loihi2, Tianjic \\
\hline
Analog architecture &	Continuous-time, low-power, biological neuron dynamics &	Low power, real-time processing, more biological realism &	BrainScaleS, Neurogrid \\
\hline	
Hybrid architecture	& Combines analog and digital computation &	Flexibility of digital, efficiency of analog &	SpiNNaker, BrainScaleS-2 \\
\hline		
\hline	
Memristive device	& Non-volatile memory-based synapses (memristors) &	In-memory computation, persistent learning & Resistive Random Access Memory (RRAM) \\
\hline		
Optical device & Uses light (photons) for spike transmission &	Extremely fast, high bandwidth, energy-efficient &	Photonic processors, silicon photonics \\
\hline
Ferroelectric device &	Uses ferroelectric materials & low power, high endurance &	Ferroelectric memory (FeRAM), ferroelectric transistor (FeFET) \\
\hline
Optoelectronic device &	Uses optical and electronic & Enhanced flexibility, speed	& Metal halide perovskite (MHP)	 \\
\hline
Magnetic device &	Uses magnetic properties & Fast, low power & magnetoresistive random access memories (MRAM) \\
\hline
\end{tabular}
\label{tab:architectures}
\caption{Key features and advantages of various neuromorphic architectures and devices.}
\end{table}

\medskip

{\bf Neuromorphic hardware and Nheuristics}: Due to their limited maturity and availability, only a few of the proposed Nheuristics have been implemented on neuromorphic hardware. Each hardware used to implement Nheuristics has unique operating characteristics, such as how fast they operate, their energy consumption and the level of resemblance to biology. The diversity of devices and materials used to implement neuromorphic hardware today offers the opportunity to customize the properties required for a given Nheuristic to solve a specific optimization problem.

Certain optimization problems demand a level of numerical precision that current neuromorphic hardware cannot achieve. In the continuous domain, solvers such as those used for quadratic programming are limited by the low-precision, fixed-point arithmetic of digital neuromorphic processors. For example, the Loihi-2 chip is constrained to 8-bit integer weight representations. Addressing large-scale optimization problems under these limitations requires innovative strategies, such as aggregating multiple synapses or leveraging scaled numerical encoding techniques (e.g., time encoding) to enhance numerical precision.

\medskip

Digital neuromorphic hardware is scalable, programmable, and relatively easy to integrate with conventional digital systems. It provide more flexibility, less costly and can handle large and complex models. Using analog components for core computations is often justified by their closer design imitation of biological systems, while also enhancing computational efficiency, high density and energy performance. Hence, analog hardware can more naturally simulate complex neural behaviors. However they are more difficult to scale and program, and their precision can be limited due to noise and variability in components. Analog/mixed-signal designs offer the potential for greater speedup and energy efficiency compared to digital systems. This approach seeks to balance the energy efficiency of analog processing with the scalability and precision of digital circuits. 

\medskip

The implementation of randomness in the hardware is an important issue for Nheuristics. A comprehensive review of advancements in the emerging field of stochastic neuromorphic electronics can be found in \cite{hamilton2014stochastic}. A fundamental challenge is how to efficiently generate the necessary stochasticity in NC hardware (e.g., CMOS, Spintronic) \cite{misra2023probabilistic}\cite{alawad2016survey}. Some promising solutions for Nheuristics are the oscillators \cite{mostafa2015}\cite{torrejon2017neuromorphic}\cite{yang201416}, chaotic \cite{lin2021review}\cite{pfeil2016effect}, and stochastic memristors \cite{al2015memristors}, which closely resembles the primary source of randomness in the brain. Among these, the stochasticity observed in biological neural networks most closely aligns with the stochastic memristor approach. Biological synapses are inherently probabilistic, exhibiting random transmission failures that range from under 10\% to over 90\%, depending on the brain region and species \cite{maass2015spike}\cite{branco2009probability}. Additionally, biological synapses can spontaneously trigger postsynaptic potentials even in the absence of a presynaptic spike.

\medskip

NC architectures are evolving rapidly. Brainchip released recently the Akida 1000. SynSense unveiled two chips: the Xylo \cite{bos2023sub}, a 28nm chip capable of modeling 1,000 neurons, and the Speck SoC, which integrates an analog Dynamic Vision Sensor (DVS) with a digital processing unit \cite{yao2024spike}. Digital chips offer varying degrees of flexibility, allowing for the modeling of different neuron types and architectures. At present, emerging neuromorphic devices (e.g., memristive, optical, ferroeletric) have not yet achieved a production-ready status and still emerging with challenges in fabrication, scaling, precision and reproducibility. However, this domain is rapidly evolving. Those various technologies of devices will provide future energy efficient, massively parallel, and very low latency.

\subsection{Hardware-aware development of Nheuristics}

\begin{figure}[h]
	\centering
	\includegraphics[scale=0.5]{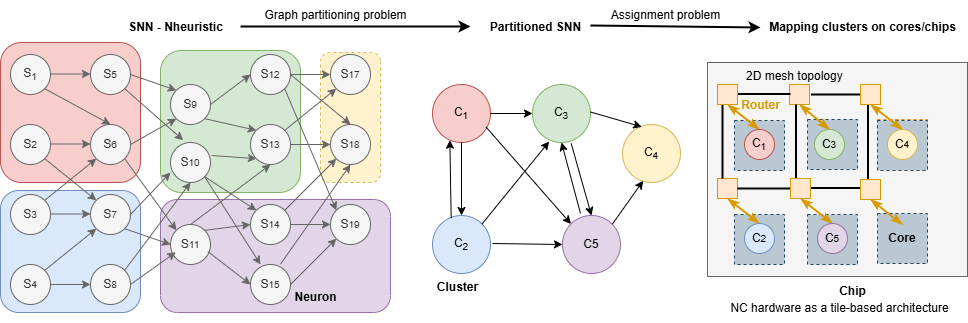}
	\caption{Mapping an Nheuristic on a given NC hardware: SNN partitioning in clusters, and cluster mapping on multiple cores of a single chip. }
	\label{fig:mapping}
\end{figure}

To the best of our knowledge, effective methods for hardware–Nheuristic co-optimization—aimed at enabling efficient, hardware-aware development of Nheuristics—have yet to be thoroughly explored by researchers \cite{huynh2022implementing}\cite{bouvier2019spiking}. Two distinct problems are identified based on the reconfigurability of the target hardware:
\begin{itemize}
\item {\bf Reconfigurable hardware optimization}: There is a substantial opportunity for co-design across the entire computing stack, enabling Nheuristics to influence the underlying hardware design (Fig. \ref{fig:codesign}) \cite{schuman2022opportunities}. This approach allows for the customization of hardware implementations tailored to specific algorithms and optimization problems. One can exploit the synergy between Nheuristics and neuromorphic hardware through their concurrent design and optimization. It not only expands the focus beyond digital computing but also promotes a re-examination of analog/mixed architectures, and emerging neuromorphic devices. 

\begin{figure}[h]
	\centering
	\includegraphics[scale=0.5]{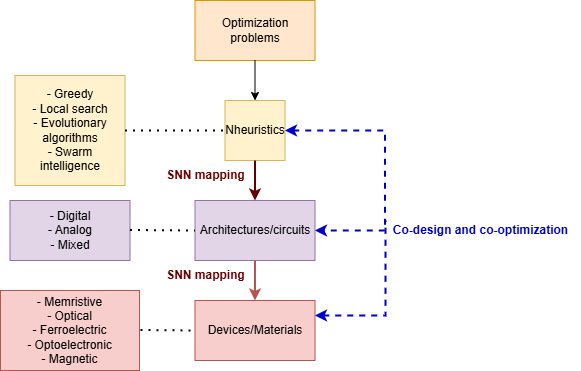}
	\caption{Hardware-aware development of Nheuristic. (a) co-design and co-optimization of hardware-Nheuristics. (b) SNN mapping on hardware.}
	\label{fig:codesign}
\end{figure}

\item {\bf SNN-hardware optimal mapping}: The approach used to map an SNN onto a given neuromorphic hardware can greatly affect its performance (Fig. \ref{fig:codesign}). As neuromorphic hardware and SNNs scale up, efficiently mapping a large SNN to the hardware becomes increasingly challenging \cite{jin2023mapping}. 

All large-scale neuromorphic hardware platforms follow a common design principle using a tile-based architecture \cite{Rajendran2019}, where tiles are interconnected through a shared communication network (Fig.\ref{fig:hardware}). Each tile typically consists of: (i) a neuromorphic core, responsible for implementing neuron and synapse circuits; (ii) peripheral logic for encoding and decoding spikes into Address Event Representation (AER); (iii) a network interface to send and receive AER packets through the interconnect. Switches are strategically placed within the interconnect to efficiently route AER packets to their destination tiles. Address-Event Representation (AER) is an event-driven neuromorphic protocol designed for inter-chip communication. Initially proposed by Sivilotti \cite{sivilotti1991wiring}, AER enables the transmission of spatial and temporal information of sparse neural events between neuromorphic chips. Table\ref{Tab:chips} presents the capacity of some recent NC hardware. 

\begin{figure}[h]
	\centering
	\includegraphics[scale=0.5]{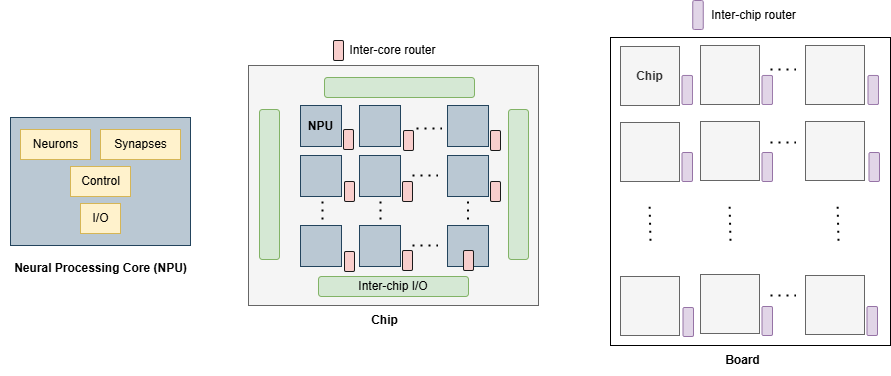}
	\caption{Description of NC hardware at different scales. A neuromorphic chip consists of multiple NPUs interconnected to facilitate communication of Address Event Representation (AER) events between cores. A neuromorphic board incorporates multiple neuromorphic chips, enabling large-scale parallel processing.}
	\label{fig:hardware}
\end{figure}

{\small
\begin{table}[h!]
\begin{tabular}{|c|c|c|c||c||c|c|}
\hline
& \# Neurons/core & \# Synapses/core  & \# Cores/chip & \# Chips/board  & \# Neurons   & \# Synapses \\
\hline
ODIN \cite{frenkel20180} & 256 & 64K & 1 & 1 & 256 & 256 \\
\hline
$\mu$Brain \cite{stuijt2021mubrain} & 336 & 38K & 1 & 1 & 336  & 336 \\
\hline
DYNAP \cite{moradi2017scalable} & 256  & 16K & 1 & 4 & 1K & 65K \\
\hline
BrainScaleS \cite{schemmel2012live} & 512 & 128K & 1 & 352 & 4M & 1B \\
\hline
SpiNNaker \cite{furber2014spinnaker} & 36K & 2.8M & 144 & 56 & 2.5B  & 200B \\
\hline
NeuroGrid \cite{benjamin2014neurogrid} & 65K & 8M & 128 & 16 & 1M & 16B \\
\hline
Loihi \cite{davies2018loihi} & 130K & 130M & 128 & 768 & 100M & 100B \\
\hline
\end{tabular}
\caption{Capacity of some NC hardware.}
\label{Tab:chips}
\end{table}
}

The efficiency of the mapping process is strongly influenced by both the SNN architecture and the hardware topology \cite{balaji2019mapping}. A typical process for mapping an SNN application to hardware (Fig. \ref{fig:mapping}) involves: (i) partitioning the neurons into multiple clusters to meet hardware constraints (e.g., limited neuron and synapse capacities, constrained neural architecture) and minimize spike communications and maximizing resource utilization, and (ii) placing these clusters onto neuromorphic computing cores for efficient execution. An important objective is to minimize the path length between routers. Each router is connected to a core and links to neighboring routers in four directions via bidirectional connections, forming a 2D mesh interconnection network (Fig. \ref{fig:mapping}). However, most existing approaches lack scalability and are limited to small to medium scale SNNs and hardware \cite{song2022dfsynthesizer}\cite{balaji2019mapping}\cite{galluppi2012hierachical}\cite{jin2023mapping}\cite{das2018mapping}\cite{ji2016neutrams}. These two NP-hard problems can be formulated as multi-objective optimization problems, as they consider multiple factors such as latency, energy consumption, cost, and reliability. To the best of our knowledge, all existing studies formulate the mapping problem using a single-objective formulation \cite{huynh2022implementing}\cite{bouvier2019spiking}.
\end{itemize}

\subsection{Simulation-based implementation of Nheuristics}


Simulating Nheuristics is critical for understanding their behavior, testing algorithms, and developing applications before implementing them on neuromorphic hardware \cite{kulkarni2021benchmarking}. Various software tools have been developed during the last decade to simulate SNNs efficiently, ranging from research-based simulators to platforms that integrate neuromorphic hardware for real-time applications. Current simulators vary in their level of biological modeling, computational speed, and compatibility with different hardware platforms. The most prominent simulation tools for SNNs are: 

\begin{itemize}

\item {\bf NEURON}: NEURON is one of the most widely used simulation environment for modeling SNNs, particularly in biological and computational neuroscience \cite{Carnevale2010}. It key features are the support of detailed modeling of both single neuron and large scale networks dynamics using Hodgkin-Huxley-type models (i.e., using ionic channels, dendrites and synaptic dynamics). NEURON is less optimized for SNNs that require high performance and real-time simulation.

\item{\bf NEST}: NEST is a highly scalable simulator for large-scale SNNs (i.e., millions of neurons and billions of synapses) \cite{diesmann2001nest}. It is particularly designed for large-scale simulations of brain regions or whole-brain models. NEST focuses on the dynamics, scale, and structure of neural networks. It includes neuron models like integrate-and-fire and models of plasticity such as STDP. NEST has an excellent scalability and performance and supports high-performance computing environments. However it is not suitable for detailed biophysical neuron models compared to NEURON.

\item {\bf Brian2}: Brian2 is a flexible, user-friendly Python simulator for SNNs, designed for easy and quick modeling while retaining powerful computational performance \cite{stimberg2019brian}. It is known for its flexibility, ease of use and extension, and frequent use in computational neuroscience. It supports both simple and complex neuron models, such as leaky integrate-and-fire (LIF) and Hodgkin-Huxley neurons. Users can define custom models using differential equations. Brian2 is adapted to medium-scale SNNs and may not scale as well as NEST.

\item {\bf BindsNET}: BindsNET is a open source Python package built on top of PyTorch deep learning library, focusing on reinforcement learning and online learning in SNNs \cite{hazan2018bindsnet}. BindsNET enables researchers to test software prototypes on CPUs or GPUs before deploying the model to specialized hardware. It combines the flexibility of PyTorch with SNNs; ideal for integrating SNNs with machine learning.

\item {\bf CARLsim}: CARLsim is a GPU-accelerated simulation environment designed for large-scale simulations of SNNs with a focus on real-time simulation and high performance \cite{niedermeier2022carlsim}. CARLsim is a user-friendly library written in C++ that is GPU-accelerated and supports CPU-GPU co-execution. It supports detailed neuron models and synaptic plasticity mechanisms, including STDP, and can simulate networks of up to a million neurons in real time.

\item {\bf Nengo}: Nengo is a versatile neural simulator that supports both spiking and non-spiking neural networks \cite{bekolay2014nengo}. It allows users to build neural models for various applications, including robotics, deep learning, and cognitive modeling. It is developed in Python and supports the TensorFlow back end. It can be deployed on neuromorphic hardware, such as Loihi and SpiNNaker, and provides tools for creating biologically plausible neural models and cognitive systems. Nengo is easy to use and versatile, but more focused on cognitive modeling and real-time applications rather than large-scale brain simulations.

\item {\bf GeNN}: GeNN is a GPU-accelerated flexible framework designed to simulate medium to large-scale SNNs efficiently \cite{knight2021pygenn}. It allows users to define SNN models in a way similar to Brian2 but runs the simulations on a GPU. GeNN is less user-firendly compared to NEST and Brian2.

\item{\bf Hermann}: Hermann is a simple and fast framework for simulating SNNs, compared to NEST and Brian2. It supports the implementation of various neuron models, with a focus on high performance and scalability, and less suitable for biologically plausible neuron models. It is designed to work well on both multi-core CPUs and GPU platforms.

\item {\bf LAVA}: LAVA is an open-source software framework that allow prototyping and programming neuromorphic applications. It offers a set of tools and abstractions for developing neuromorphic applications on hardware such as Intel Loihi. It allows also prototyping applications on CPU and GPU.

\item{\bf PyNN}: PyNN is a Python-based cross-platform framework designed to provide a unified interface for several SNN simulators, including NEST, NEURON, and Brian2 \cite{davison2009pynn}. It provides a standard interface for different simulators, allowing flexibility and portability. PyNN supports multiple SNN models and synaptic plasticity rules. It depends on external simulators for actual computation and is not optimized for real-time or large-scale simulation itself. Moreover, PyNN supports mapping SNN models on NC hardware such as SpiNNaker, BrainScaleS, and Loihi.

\item{\bf SpikingJelly}: SpikingJelly is an open-source framework, which bridge SNNs and deep learning \cite{fang2023spikingjelly}. It supports the simulation of SNNs on both CPUs and GPUs.

\end{itemize}

\begin{table}[h!]
\begin{tabular}{|c|c|c|c|c|c|c|c|}
		\hline
		Software & User  & Flexibility & GPU  & NC Hardware  & Biologically  & Performance & Scalability \\
		 &  friendly &  & Handling & support &  plausible &  &  \\
		\hline
		NEURON & ++ & ++ & No & No & +++ & ++ & + \\
		\hline
		\href{https://www.nest-simulator.org/}{NEST} & ++ & ++ & No & No & +++ & +++ & +++ \\
		\hline
		\href{https://brian2.readthedocs.io/en/stable/}{Brian2} & +++ & ++ & No & No & +++ & ++ & + \\ 
		\hline
		\href{https://bindsnet-docs.readthedocs.io/}{Bindsnet} & +++ & ++ & Yes & No & ++ & + & ++ \\ 
		\hline
		\href{https://uci-carl.github.io/CARLsim3/}{CARLsim} & ++  & ++ & Yes &  No  & ++ & +++ & +++ \\
		\hline
		\href{https://www.nengo.ai/}{Nengo} & +++ & +++ & Yes & Yes & ++ & +++ & ++ \\ 
		\hline
		\href{https://genn-team.github.io/}{GeNN} & + & ++ & Yes & No & +++ & +++ & +++ \\
		\hline
		Hermann & +++  & ++  & Yes & No & + & +++ & +++ \\
		\hline
		\href{https://lava-nc.org/}{Lava}  & ++ & +++ & Yes & Yes & ++ & ++ & ++ \\
		\hline
		\href{https://neuralensemble.org/PyNN/}{PyNN} & ++ & +++ & Yes & Yes & ++ & ++ & +++ \\
		\hline
		\href{https://spikingjelly.readthedocs.io/zh-cn/latest/}{Spiking Jelly} & +++ & ++ & Yes & No & ++ & ++ & ++ \\ 
		\hline
\end{tabular}
\caption{Multi-criteria comparison of simulation software tools.}
\label{Table:simulation}
\end{table}

\medskip

{\bf Simulators and Nheuristics}: Choosing a simulator can be challenging for newcomers in the SNN community. Some existing simulators, such as Brian2, NEURON, NEST, and GeNN, primarily address computational neuroscience issues. However, their performance and flexibility are generally insufficient for use in Nheuristics. Other simulators, such as BindsNet, Nengo, SpikingJelly, and CARLsim, are designed for machine learning applications and are integrated with deep learning frameworks like PyTorch and TensorFlow.

\medskip

Performance, scalability, and flexibility (i.e., the capability to tackle a wide range of tasks) are key factors to consider when selecting a simulator for Nheuristics. Currently, the most suitable simulators for this purpose are LAVA and PyNN. Both offer ease of use, flexibility, and scalability, along with support for neuromorphic hardware and are platform-agnostic.

\section{Classification and analysis of Nheuristics}
\label{sec:neuro-meta}

Most existing Nheuristics are designed for specific optimization problems, with only a few being general enough to be applied across a variety of optimization problems, including COPs and DOPs. Due to the inherent information encoding and network structure of SNNs, optimization problems that align well with SNNs are typically BOPs and graph-related problems. The spike-based nature of SNNs—relying on the presence or absence of spikes—makes them particularly well-suited for solving BOPs and graph problems, such as SAT, QUBO, and graph partitioning. 

\medskip


\subsection{Computation of the objective function}

A key challenge in Nheuristics is the computation of the objective function. If a Von Neumann processor is required to evaluate the objective function, it creates an impractical off-chip communication bottleneck that worsens as the problem size increases. It is much more efficient to compute the objective function in a distributed, event-driven manner within the SNN, communicating with the CPU only when a solution is found. As a result, most existing work focuses on optimization problems where the objective function is easily computed within the NC paradigm (e.g., matrix-vector products in QUBO, free calculations for 3-SAT). However, a major obstacle to the further advancement of Nheuristics is the computation of complex objective functions, particularly those involving non-linear functions. Three alternatives can be investigated:
\begin{itemize}
\item {\bf Numerical SNN building blocks}: Numerical solutions are a natural fit for the Von Neumann computation paradigm. SNNs approaches have been often considered non adapted for high-precision arithmetics \cite{aimone2022review}. In \cite{schuman2021neuromorphic}, the authors prove that NC is Turing-complete, making it suitable for general-purpose computing. In \cite{maass1996lower}, the author show that a deterministic SNN has the theoretical capability to be as powerful as a universal Turing machine. It can encode and transmit information using the durations of interspike intervals. Recently, NC have shown promising results in traditional scientific computing applications such as matrix operations \cite{parekh2018constant}, partial differential equations solvers \cite{smith2020solving}, utility functions (e.g., sorting) \cite{verzi2018computing}. The development of SNN building blocks, with several useful kernels will facilitate the use of NC for numerical applications.

\item {\bf Approximated SNNs}: NC can be employed to approximate the function to optimize. Here, a SNN is trained to reproduce the behavior of the objective function. This approach is also adapted to solve optimization problems characterized by expensive objective functions (e.g., black box simulation). A scientific challenge is what type of SNN architecture can be designed to provide general and accurate approximation capabilities \cite{maass1997fast}\cite{iannella2001spiking}.

\item {\bf Heterogeneous NC-digital models}: one can use an heterogenous platform combining NC and Von Neumann paradigms. Here, the objective function can be computed on a digital computer (e.g., GPU, CPU, multi-core, or even a parallel hetergenous CPU-GPU architecture) \cite{nilsson2023integration}. Using NC, it is possible to halt all neuromorphic activity at any time if necessary. This contrasts with quantum computing algorithms, where measuring the quantum system destroys the state of the qubits. Hence, communication between NC programs with classical processors (CPUs or GPUs and their memory) remains possible even when a neuromorphic program is paused or stopped. However, the disadvantage of such alternative is the power consumption, latency and footprint of the metaheuristic using the Von Neumann paradigm.
\end{itemize}

\subsection{Energy consumption}

Measuring the energy consumption of Nheuristics is an important issue. Indeed, balancing performance with sustainability is important for the future of computing growth. Power consumption is mainly driven by the propagation of spikes, which are the primary factors behind data movement and processing. Measuring the power consumption of a Nheuristic on specific hardware is straightforward. It allows for comparisons of the power usage of the same Nheuristic across different hardware platforms. In \cite{pierro2024solving}, the authors present initial performance comparisons of a neuromorphic SA against SA running on a CPU. The results demonstrate that the Loihi 2-based SA implementation can find feasible solutions for problems with up to 1000 variables in just 1 ms, while consuming 37 times less power than the CPU baseline algorithms. In \cite{alom2017quadratic}, a Nheuristic designed to solve the QUBO problem consumes 50mW of power. The authors also note that around 50\% of the total power consumption in TrueNorth is passive power.

\medskip

It is also important to define a energy models to estimate the energy consumption of SNNs, independent of any specific hardware. An energy model allows for predicting a network's energy expenditure on a target system without the need for direct access to it. Most energy consumption metrics are based on the number of synaptic operations, such as accumulations (ACC), in the SNN \cite{barchid2023spiking}. These metrics have limitations, as they equate energy consumption solely with the energy used by synaptic operations \cite{lemaire2022analytical}. In \cite{ostrau2022benchmarking}, a more realistic energy model is proposed to estimate the power consumption of a network simulated on a target system. This model considers the static power consumption of the idle neuromorphic system, the energy required for a virtual spike source neuron to emit an event, the power needed to simulate or emulate an idle neuron, the energy used by a real neuron to emit a spike, the energy spent transmitting a single spike, and the static power required to activate STDP, along with the energy per synaptic event.

\medskip

In the following subsections, we examine the key families of Nheuristics, namely greedy algorithms, local search, swarm intelligence, and evolutionary algorithms.

\subsection{Greedy Nheuristics}
\label{sec:greedy}

Greedy Nheuristics represent natural candidates for some graph problems such as shortest path problems \cite{davies2018loihi}, neighborhood subgraph extraction \cite{schuman2019shortest} and minimum spanning trees (MST) \cite{kay2020neuromorphic}. First, a graph-SNN embedding is carried out. Synaptic delays can be defined by edge weights. Then, neurons are selectively stimulated, and the SNN is executed to simulate activity within the network. The resulting spiking activity or updated network—altered through mechanisms like synaptic plasticity—provides a solution to the graph problem. In \cite{davies2018loihi}, an SNN for finding the single-source shortest path is presented, leveraging the event-driven characteristics of SNNs and synaptic plasticity to identify the shortest routes between nodes. In \cite{schuman2019shortest} the shortest path and neighborhood subgraph extraction have been solved on memristive devices. Each synapse is realized using two memristors, enabling the representation of both positive and negative weight values. The SNN is defined by a set of LIF neurons and STDP learning. The set of synapses have programmable delays which are equal to the weights of the graph. By setting the refractory period to be longer than the sum of the edge lengths in the graph, one can ensure that each neuron in the network fires no more than once. Hence, the path length from the source node to every node in the graph is determined by examining the firing times of each neuron. In \cite{kay2021neuromorphic}, the authors provide a preprocessing technique which produces an SNN whose topology is the same as the graph, but with different synaptic delays. This strategy prevents the issue of stimulating a single vertex causing multiple neurons to fire at the same time. 

\medskip

An essential component for path planning and navigation in nature are {\it Place Cells}, found in the hippocampus and discovered by O’Keefe in 1971 \cite{o1971hippocampus}. Hence, some greedy Nheuristics draw inspiration from spike wavefronts observed in the hippocampus during route planning. These wavefronts seem to play a role in the mental search for the shortest path through a spatial map \cite{dragoi2013distinct}. Ponulak and Hopfield modeled how the brain could solve spatial navigation problems by simultaneously exploring alternative routes through propagating waves of spiking activity \cite{ponulak2013rapid}. As the wavefront progresses, synapses are modified according to plasticity rules, enabling the path to be retrieved later from the network. 

\medskip

This approach has been applied to some real-life path planning problems. In \cite{chao2023brain}\cite{koul2019waypoint}\cite{ruan2024gsnn}\cite{zennir2017robust}, the authors propose a drone/robot path planner. The topology of the SNN corresponds to a discretization of the environment. The SNN can be viewed as an interconnected network of place cells that creates a cognitive map, serving as a medium for a propagating wave that moves through the SNN with the sequential activation of different neurons. The solution is represented as a synaptic vector field, while obstacles are represented by inhibited neurons.  The firing rate of the place cells increases as they get closer to the current position of the drone/robot.

\subsection{Local search based Nheuristics}
\label{ls}

Local search based Nheuristics (LS-Nheuristics) are among the most popular Nheuristics for solving small size instances of binary, discrete, and continuous optimization problems. 

\medskip

{\bf LS-Nheuristics for binary optimization problems}: In general, LS-Nheuristics are influenced by the pioneering work of Hopfield and Tank \cite{hopfield1986computing} on BOPs using Boltzmann machines. Most existing LS-Nheuristics are stochastic and have been developed for QUBO. When addressing the QUBO problem, the SNN topology is typically aligned with the input graph of the QUBO instance.

\medskip

In stochastic LS-Nheuristics, the spiking of a stochastic noise is introduced into the SNN as a computational resource for the generation of new solutions. In \cite{alom2017quadratic}, a stochastic LS-Nheuristic is implemented on the TrueNorth hardware using the IF neuron model and a recurrent SNN architecture. It consists to generate from $X_n$ the new solution $X_{n+1} = \{x / \sum_{y \in X} Q_{x, y} \geq 1 \}$ and repeat this step until a given number of ticks\footnote{The tick is the duration of the neuron and synapse update. In TrueNorth, the tick is equal to a millisecond.}. Using LIF neurons coupled with random decaying noise generators, it is possible to get a fast convergence towards a very good approximate solution, by reproducing this scheme with stochastic properties. The LIF model can be described by the following equations:
\begin{equation}
	\frac{dV_j}{dt} = \sum_{0 \leq i < D} \mathbbm{1}_{i \in X} \times Q_{i, j} -  \alpha V_j
\end{equation}
\begin{equation}
	V_j(t + \delta t) = \left\lbrace \begin{aligned}
		& V_j(t) + \frac{dV_j}{dt} \text{ if } V_j(t) < V_{th} \\
		& 0  \text{ and Spike} \text{ if } V_j(t) \geq V_{th}
	\end{aligned} \right.
\end{equation}
where $\alpha$ represent the leaky term, $V_{th}$ represents the membrane threshold and $X$ is updated to contain every spiking neuron at time $t$. To take the noise into account, simply add to $X$ the neurons that respects $	\mu \beta^t \leq R_i, \forall i \in \{0, \dots, n-1\}$, where $R_i$ is a random variable chosen in $[0, 1]$ with the uniform distribution, $\mu$ is the initial probability of spiking and $\beta$ the decaying factor.

\medskip

A similar approach has been extended for SA in \cite{mniszewski2019graph}. A stochastic leak has been introduced to serve as the temperature parameter in SA. In \cite{pierro2024solving}, the authors implement a SA on the Loihi2 neuromorphic processor. To speed up the convergence, they also enable parallel updates of many neurons, which pushes the Boltzmann machine away from a near-equilibrium state where neurons flip one at a time. They introduce a stochastic refractory period, preventing neurons from repeatedly flipping variables in successive steps. Specifically, after a variable neuron changes its state, from 0 to 1 or vice-versa, further changes are inhibited for a random number of iterations\footnote{It can be seen as a tabu search strategy.}. In \cite{henke2023sampling}, the authors show that Loihi2 implementation of SA outperforms a standard linear-schedule anneal on a D-Wave quantum annealer for the QUBO problem. The authors also introduce an iterated SA where the best solution found is used to initialize the next run of SA. Some LS-Nheuristics have been implemented on emerging neuromophic devices. In \cite{yin2024ferroelectric} (resp. \cite{cai2020power} \cite{yang2020transiently}, \cite{shin2018hardware} \cite{mahmoodi2019analog}), a SA has been mapped on a ferroelectric field effect transistor (FeFET) device (resp. Memristor, RRAM), which allows to accelerate vector-matrix multiplications of QUBO formulations. Some LS-Nheuristics have also been explored for other BOPs. In \cite{yakopcic2020solving}, a LS-Nheuristic based on a stochastic neuron model was proposed to solve SAT problems on the Loihi processor.

\medskip

\begin{figure}
	\centering
	\includegraphics[scale=0.5]{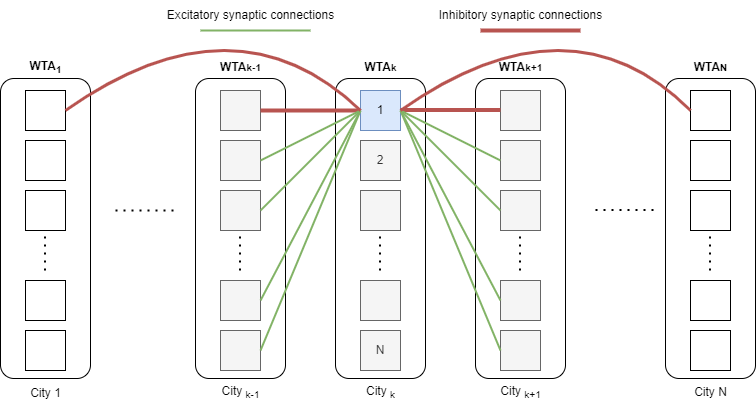}
	\caption{SNN solving the TSP problem using WTA subnetworks.}
	\label{fig:illustrative}
\end{figure}

{\bf LS-Nheuristics for discrete optimization problems}: LS-Nheuristics have been mainly used to solve the popular CSP and TSP problems. Winner-takes-all (WTA) networks represent an important SNN architecture to solve DOPs \cite{maass2000computational}. In solving the TSP of size $N$, the SNN can be constructed as follows. A separate WTA subnetwork \( WTA_k \) is dedicated to each step \(  k\) of the tour. Each WTA subnetwork consists of \( N \) neurons, one for each of the \( N \) cities that could potentially be visited at step \( k \). Strong negative weights between the \( N \) neurons within each WTA subnetwork ensure that, typically, only one neuron fires at any given moment. The index \( i \) of the neuron in WTA module \( WTA_k \) that most recently fired can be interpreted as the proposed city \( i \) to visit at step \( k \) of the tour. Therefore, by recording which neuron in each WTA subnetwork has fired most recently, the firing activity of the entire network can be decoded at any time \( t \) to propose a TSP solution. In Fig.\ref{fig:illustrative}, the green links between adjacent WTA modules, which select cities for consecutive steps in the tour, represent excitatory synaptic connections. The strength (weight) of these connections encodes the distance matrix between the \( N \) cities. Weights are larger for synaptic connections between neurons encoding cities \( i \) and \( j \) that are close to each other, increasing the likelihood that if one neuron fires, the other will also fire, thus proposing a direct route from city \( i \) to city \( j \) (or vice versa). Strongly negative weights on synaptic connections, represented by the red arcs, between neurons with the same index \( i \) in different WTA subnetworks, reduce the probability of generating a tour in which city \( i \) is visited more than once.

\medskip

\begin{figure}
	\centering
	\includegraphics[scale=0.55]{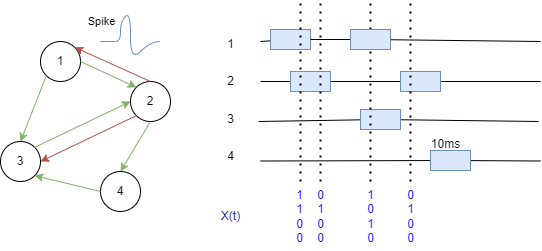}
	\caption{The network state $x(t)$ can be read at any time $t$ from the network activity. A neuron state $x(t)$ cannot be changed in the period $[t-\tau,t]$} (e.g., $\tau=10ms$.)
	\label{fig:barriers}
\end{figure}

Designing the weights in a manner that ensures the best tours have the highest likelihood of being selected is a challenging task. In \cite{buesing2011neural}, the authors provided a clear relationship between the weights and biases of neurons and the probability of producing a specific TSP tour, i.e., a particular network state \( \{x_1, \ldots, x_N\} \). The probability of a given state \( \{x_1, \ldots, x_n\} \) at time \( t \) is expressed as:
\[
p(\{x_1, \ldots, x_N\}) = \frac{1}{z} \exp\left(\frac{-E(\{x_1, \ldots, x_n\})}{T}\right)
\]
where \( z \) is a normalization factor, $T$ controls the degree of stochasticity, and the energy function \( E(\{x_1, \ldots, x_N\}) \) is defined as:
\[
E(\{x_1, \ldots, x_n\}) = \sum_{i,j \in G} w_{ij} x_i x_j - \sum_i b_i x_i
\]
This formula makes it straightforward to define the weights \( w_{ij} \) and biases \( b_i \) in a SNN such that TSP tours that satisfy all constraints and have the smallest total distance will have the lowest energy, and thus the highest probability in the stationary distribution of this Markov chain. The energy term \( E(\{x_1, \ldots, x_N\}) \) is commonly referred to as the "energy" of the state, and the network tends to move towards states that minimize this energy. The probability \( p(\{x_1, \dots, x_N\}) \) of an SNN state \( \{x_1, \dots, x_N\} \) is the same as that of a commonly used stochastic ANN, the Boltzmann machine. The fine-scale timing dynamics of SNNs enable them to efficiently escape from local minima, making them more effective at finding the global minimum compared to Boltzmann machines, despite both sampling from the same underlying distribution. Indeed, Boltzmann machine tends to remain in each visited state for a longer duration. When neuron $i$ fires, the corresponding component $x_i$ of the state vector is set to $1$ for a duration $\tau$. This can be interpreted as the equivalent of the tabu list size in the well-known tabu search metaheuristic \cite{glover1990tabu}. After this period, the neuron is automatically resets to $0$, regardless of the energy difference caused by this transition. This mechanism leads to a high frequency of state changes that cross significant energy barriers. In contrast, a Boltzmann machine determines each state transition based on the energy difference between states, thereby avoiding transitions that would increase the network's energy (Fig.\ref{fig:barriers}). If the network contains a sufficient level of noise, such as that introduced by the stochastic firing of neurons, the distribution of network states rapidly converges—exponentially fast—from any initial state to a unique stationary (equilibrium) distribution $p(x)$ \cite{habenschuss2013stochastic}.

\medskip

Similarly, in \cite{jonke2016solving}, a stochastic neuron model \cite{jolivet2006predicting} and timing-based information encoding were applied to solve the TSP and 3-SAT problems. These studies highlight how stochastic spike timing dynamics can enhance the computational capabilities of SNNs, resulting in highly efficient and rapid problem-solving.

\medskip

\begin{figure}[h]
	\centering
	\includegraphics[scale=0.55]{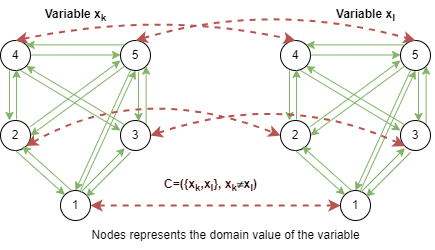}
	\caption{WTA subnetwork encoding an all different constraint between two variables $x_k$ and $x_l$. The domain of each variable has 5 possible values. Red (resp. green) edges represent inhibitors (resp. excitatory) synaptic connections}
	\label{fig:wta-CSP}
\end{figure}

The most extensively studied DOP for LS-Nheuristics is the CSP. 
Most of the proposed LS-Nheuristics are inspired by the seminal work of Hopfield and Tank on solving CSPs with Boltzmann machines \cite{hopfield1986computing}. The stochastic SNNs governed by an energy function \cite{davies2021advancing}\cite{aarts1989simulated}:
$$
E = S^T(t).W.S(t) = \sum_{i}{(S_i. \sum_{j}{W_{ij}.S_j})}
$$
to solve QUBO problems, where $S$ is the instantaneous spike vector and $W$ the synaptic weight matrix. The SNN is structured such that different values of the CSP variables $X$ are represented by one-hot coded winner-takes-all (WTA) subnetworks, while $W$ encodes the constraints $C$. Figure \ref{fig:wta-CSP} presents an illustrative example of modeling an all-different constraint using an SNN. Each possible value of a variable is represented by a neuron. Inhibitory synapses (red edges) enforce the constraint by connecting neurons representing the same value, while excitatory synapses (green edges) allow value changes for a given variable. The resulting SNN forms a dynamical system for solving the CSP. Various strategies can then be implemented to introduce stochastic noise, enabling exploration of the search space to find a solution that satisfies all constraints. In \cite{fonseca2017using}, a spiking stochastic noise is leveraged as a computational resource to facilitate search space exploration. This noise mechanism enables broader sampling (by adding or removing variables) through stochastic leak/decay processes. The probability of firing or leakage acts as the temperature component in simulated annealing (SA), where the initial temperature is set high and gradually decreases according to an annealing schedule. 

\medskip

LS-Nheuristics have been investigated on different hardware. In \cite{guerra2020stochastic}, an SNN was implemented on SpiNNaker and Loihi using a recurrent network with a LIF neuron model and winner-take-all (WTA) learning to solve the vertex coloring problem. In \cite{liang2019neuromorphic} (resp. \cite{hong2021memory}), a stochastic LS-Nheuristics has been implemented on the DYNAP hardware \cite{moradi2017scalable} (resp. RRAM). The SNN is composed of two main sub-networks: the variable-encoding networks and the constraints networks. Each variable-encoding network represents a variable in the CSPs, with the variable's value indicated by a winning population of neurons in a Winner-Takes-All (WTA) network. WTA winners can also signify meta-stable states, indicating that the variable has not yet been assigned a value. The constraint networks define the relationships (constraints) between variables. All variable and constraint networks operate simultaneously, working together to determine the final equilibrium state of the system. In \cite{binas2016spiking} a stochastic LS-Nheuristic solving the CSP has been implemented on a reconfigurable analog circuits, which implements thermal noise and adaptive exponential IF neurons models. The device’s intrinsic noise has been used as a source of randomness. In \cite{corder2018solving}, the authors demonstrate how the vertex cover problem can be solved on TrueNorth using the Ising spin model \cite{zhang2024review}. The Metropolis-Hastings algorithm, a widely used Markov Chain Monte Carlo (MCMC) method, enables the approximation of high-dimensional distributions through iterative sampling. This Ising model is then mapped onto a network of leaky integrate-and-fire (LIF) neurons, with each node in the model represented by a small, fixed-size network of neurons. Within the Metropolis-Hastings MCMC framework, state transitions occur stochastically: lower-energy states are always accepted, while higher-energy states are accepted with a probability dictated by the Boltzmann distribution.

\medskip

\begin{figure}[h]
	\centering
	\includegraphics[scale=0.55]{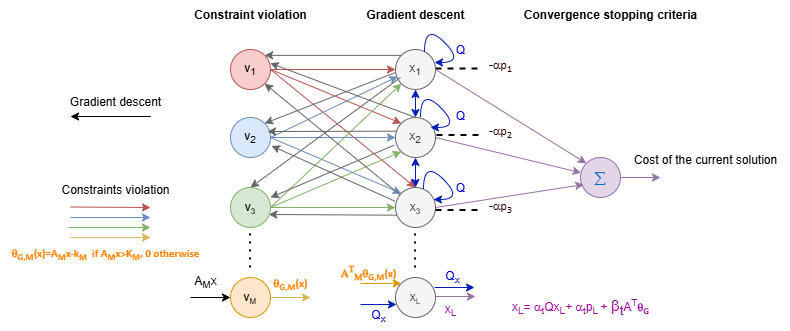}
	\caption{A gradient descent for Quadratic Programming (QP) with linear constraints. Each grey circle represents a neuron performing gradient descent updates, while dashed lines indicate biases. Colored circles represent neurons that compute constraint violations and apply corrections for each constraining plane. A purple blue integration neuron monitors the cost of the current network state. $\theta_G(x_t)$ represents the rectified linear unit function. The hyperparameters $\alpha_t$ gradually decreases, while $\beta_t$ increases over time as the SNN converges toward the minimum of the search space.}
	\label{fig:QP}
\end{figure}

{\bf LS-Nheuristics for continuous optimization problems}: Very few LS-Nheuristics has been investigated for COPs \cite{davies2021advancing}. The development of the spiking Locally Competitive Algorithm  (LCA) \cite{tang2017sparse}, designed to solve LASSO, marked the first approach to solving unconstrained convex QPs with an SNN. In \cite{mangalore2024neuromorphic}, the authors present a LS-Nheuristic to solve convex COPs with quadratic cost functions and linear constraints (QP) on Loihi 2 chip. A two-layer recurrent SNN has been designed to minimizes the constrained cost function using gradient descent (Fig.\ref{fig:QP}). By relying solely on element-wise or matrix-vector arithmetic, this type of network can be efficiently deployed on a neuromorphic hardware.

\subsection{Neuromorphic-based swarm intelligence algorithms}
\label{si}

Swarm intelligence based Nheuristics (SI-Nheuristics), such as PSO and ACO, have been explored within the SNN framework to address both DOPs and COPs.

\medskip

\begin{figure}[h]
\centering
\includegraphics[scale=0.5]{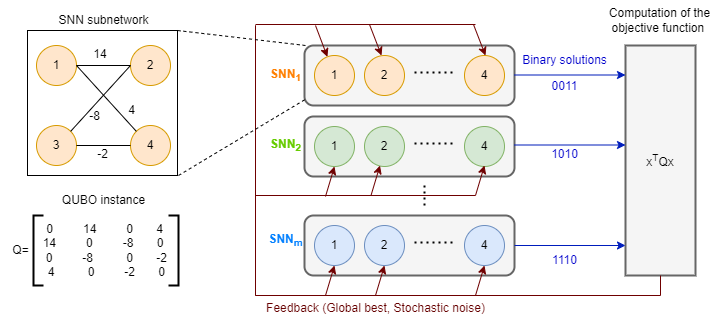}
\caption{Swarm of SNNs to solve the QUBO problem. An instance of the QUBO problem, its associated SNN structure, and the swarm of collaborating SNNs.}
\label{fig:swarm-binary}
\end{figure}

{\bf SI-Nheuristics for binary optimization problems:} Most of the SI-Nheuristics have been applied to QUBO problems. In \cite{fang2022solving}, the authors employs multiple collaborative SNNs (Fig.\ref{fig:swarm-binary}). Each SNN is composed of LIF neurons and performs a local stochastic search. A set of SNNs operate with the same dynamics but different initializations, defined by the stochastic noise. They periodically shares the global best solutions. The interaction within the SNN occurs when a new best solution is generated. By sending spikes according to the global best solution each time it is updated, the algorithm can solve QUBO much faster than multiple non-interacting networks. However, mapping a graph problem in a SNN in such a way is very specific to the QUBO problem, and this method does not provide a general framework for solving optimization problems, rather a highly-efficient QUBO-heuristic problem. Through experiments on benchmark problems, the proposed approach proves to be efficient and effective in finding QUBO optima. Specifically, it achieves speedups of 10x and 15-20x compared to the multi-SNN solver without collaboration and the single-SNN solver, respectively. A similar SI-Nheuristic has been proposed in \cite{lele2023neuromorphic}. The recurrent swarm of SNNs has been implemented on 40nm 25mW Resistive RAM (RRAM) Compute-in-Memory (CIM) SoC with a 2.25MB RRAM-based accelerator and an embedded Cortex M3 core. The $m$ SNNs are parallel and homogeneous. 

\medskip

{\bf SI-Nheuristics for discrete optimization problems:} In \cite{fang2019swarm}, the authors develop a swarm model of collaborative agents that implements ACO based SI-Nheuristics to solve the TSP. Each agent is represented by an SNN consisting of $n$ fully connected neurons, forming a WTA network. The order of spikes from the neurons encodes the solution of a single agent (ant). Synaptic weights are updated in real-time based on the spike activity and shared across multiple SNNs, mimicking the pheromone trails in ACO. The neuron model used is Izhikevich’s model, which supports various firing patterns: FS (Fast Spiking), LTS (Low-Threshold Spiking), and RS (Regular Spiking), while the encoding is based on rate firing. Furthermore, the SNN has been deployed on an emerging hardware platform, specifically the ferroelectric field-effect transistor (FeFET). The challenge in designing the dynamics of each SNN lies in ensuring that each neuron fires only once and follows the correct order during a single trip. In previous work \cite{jonke2016solving}, multiple WTA SNNs were utilized to represent the travel path of a single trip. By leveraging the inhibitory and excitatory interfaces of FeFET spiking neurons, one can represent the travel path of an individual agent using the spike train of a single SNN.

\medskip

\begin{figure}[h]
	\centering
	\includegraphics[scale=0.5]{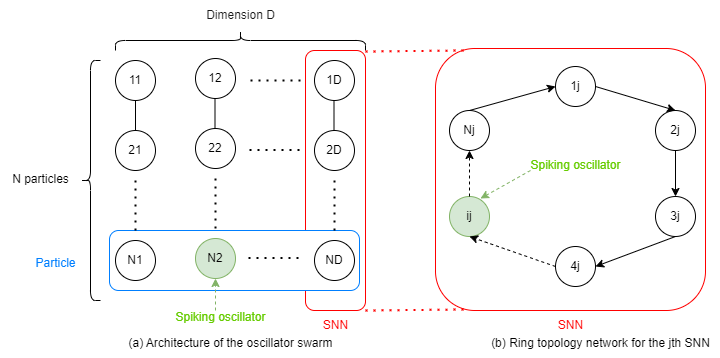}
	\caption{Architecture of the OSNN PSO Nheuristic.}
	\label{fig:PSO-continuous}
\end{figure}

{\bf SI-Nheuristics for continuous optimization problems}: Very few SI-Nheuristics have been investigated for unconstrained DOPs, $min_{x \in \R^D} f(x)$. In \cite{sasaki2023swarm}, the authors present a study on a deterministic SI-Nheuristic based on PSO, called the Optimizer based on Spiking Neural-Oscillator Networks (OSNNs). OSNNs feature a swarm composed of multiple particles that collaboratively explore a solution space by interacting with each other. Each particle is made up of $D$ spiking neural oscillators, or ``spiking oscillators'' modeled by IF neuron models (Fig.\ref{fig:PSO-continuous}). These oscillators are connected by a network topology (e.g., one-way ring) and communicate by exchanging spike signals. The interaction between coupled spiking oscillators can result in synchronous or asynchronous dynamics, which influences the search performance of OSNNs.

\medskip

The following equation is carried out by the $D$ oscillators per particle of the SNN \cite{sasaki2023swarm}\cite{yamanaka2013analysis}:
\begin{equation}
	\left[\begin{aligned}
		& y_{t+1} \\
		& v_{t+1}
	\end{aligned} \right]
	= \left\{\begin{aligned}
		& \left[\begin{aligned}
			& Q_t \\
			v_t - (&y_t - Q_t)
		\end{aligned} \right] \text{ and Spike} \text{ if } y_t \in \Pi_t \\
		& \left[\begin{aligned}
			& Q_t \\
			v_t - (&y_t - Q_t)
		\end{aligned}  \right] \text{ if Spike received} \\
		& \delta R_{\theta}
		\left[\begin{aligned}
			& y_t \\
			& v_t
		\end{aligned} \right] \text{ if } y_t \notin \Pi_t
	\end{aligned} \right.
\end{equation}

\begin{equation}
	x_t = y_t + \frac{pb_t + gb_t}{2}
\end{equation}

\begin{equation}
	\begin{aligned}
		&Q_t = pb_t - \frac{pb_t + gb_t}{2} \\
		&\Pi_t = \{y / |y| \geq |pb_t - gb_t|\}
	\end{aligned}
\end{equation}

where $pb$ and $gb$ represent the personal and global bests, respectively, $R_\theta$ is the rotation matrix with angle $\theta$, $\delta$ is a damping factor, and $\Pi_t$ denotes the spiking condition. In this setup, each neuron from dimension $i$ is connected to neurons from other particles in the same dimension, arranged in a one-way ring configuration. The oscillation can be described as a spiral movement within the phase space $(v, y)$, which resets every time $y$ exceeds a threshold relative to the distance between $pb$ and $gb$. The objective is to emulate the particle swarm search process by exploiting the unique properties of these oscillators.

\begin{minipage}{0.5\textwidth}
	\begin{figure}[H]
		\includegraphics[scale=0.2]{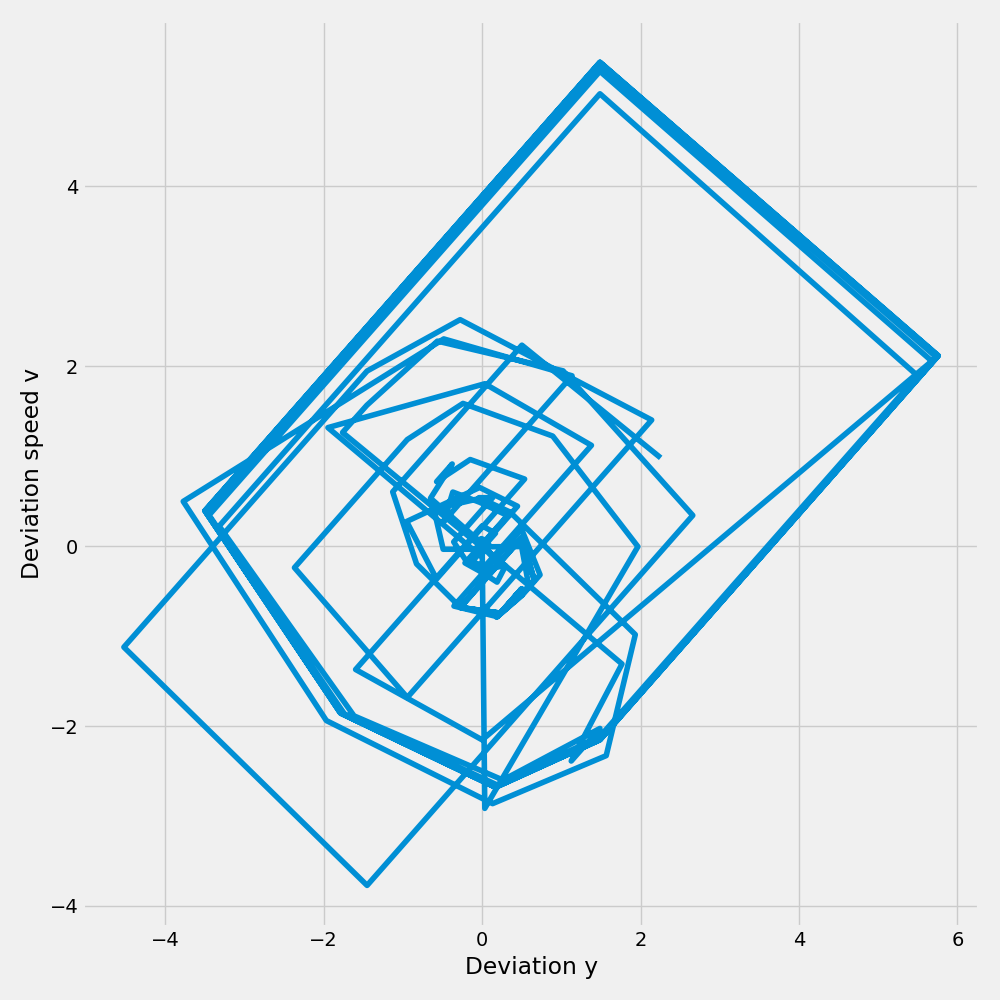}
		\caption{Visualisation of oscillations of a single OSNN during a search. $v$ given $y$ for all $t$ during the optimization of the Weierstrass's function ($N = 10, D = 7$).}
		\label{fig:PSO1}
	\end{figure}
\end{minipage}
\hspace{4ex} 
\begin{minipage}{0.5\textwidth}
	\begin{figure}[H]
		\includegraphics[scale=0.2]{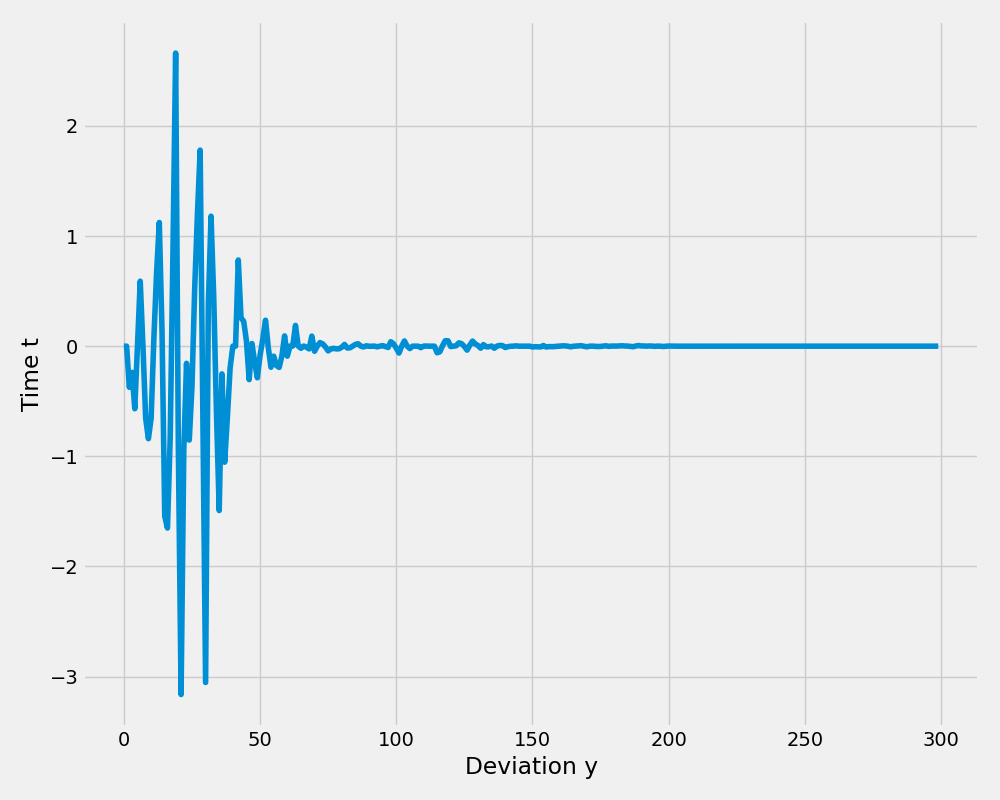}
		\caption{Visualisation of oscillations of a single OSNN during a search. $v$ given $t$ during the optimization of Levy's function ($N = 10, D = 15$)}
		\label{fig:PSO2}
	\end{figure}
\end{minipage}

\medskip

The spiking activity of neurons is sufficient to coordinate all neurons within the same dimension over time, enabling the system to converge toward a local optimum. The parameters $\theta$ and $\delta$ should be selected based on the specific problem, and the results from classical benchmarks are notably promising. However, one limitation of this approach is that the personal best ($pb$) must be shared by each particle with its respective neurons, while the global best ($gb$) needs to be shared globally. Although this is not inherently unnatural for SNNs, it can be partially mitigated by sharing this information locally and allowing it to propagate through the network, for example, by using a two-way ring to connect each particle. It is important to note that using continuous spikes would be highly energy-inefficient.
	
\subsection{Neuromorphic-based evolutionary algorithms}
\label{ea}

A pioneering approach to Nheuristics based on EAs (EA-Nheuristics) was introduced in \cite{talbi2025NEVA}. The most suitable parallel EA framework for this purpose is the cellular genetic algorithm (CEA) \cite{muhlenbein1992parallel}\cite{talbi1991CGA}. A neuromorphic evolutionary algorithm (NEVA) was proposed, structuring the CEA as SNNs. To the best of our knowledge, this represents the first evolutionary algorithm explicitly designed using the SNN computational paradigm. The algorithm leverages synaptic sparsity by employing low-degree interaction graphs, which help maintain spatial separation of local optima. Additionally, information exchange is restricted to local interactions between neurons. Computational experiments on QUBO, 3-SAT, and knapsack problems demonstrate the effectiveness of the proposed neuromorphic EA. Further improvements in both solution quality and computational efficiency were achieved by developing a neuromorphic memetic algorithm that integrates EAs with local search.

\medskip

In \cite{cruz2025}, the authors introduce the first general-purpose framework that integrates spiking neural dynamics with metaheuristics—particularly population-based approaches—for solving large-scale continuous optimization problems. The framework is built on a distributed architecture composed of Neuromorphic Heuristic Units (NHUs), each modeled using biologically inspired spiking neurons. These units encode candidate solutions and apply adaptive, spike-triggered perturbations. Signal coordination across the network is handled by a tensor contraction layer, which facilitates structured communication based on a predefined adjacency topology. A neighborhood manager regulates local interactions by directing the flow of positional data, while a high-level selector oversees global coordination by identifying and disseminating top-performing solutions. The entire system is implemented using the LAVA framework.

The table~\ref{tab:classification} below outlines the key properties of some presented Nheuristics, including their genericity, target optimization problem, neuron model, information encoding, SNN topology, learning mechanism, and implementation.

\begin{table}[h]
\begin{tabular}{|c|c|c|c|c|c|c|c|c|}
\hline
 Paper & Algo. & Type & Problem & Neuron model & Encoding & Topology & Learning & Impl. \\
\hline 
\cite{schuman2019shortest} & Greedy & Specific & Shortest path & LIF & Rate & Graph & STDP & Memristor \\
\hline 
\cite{ruan2024gsnn} & Greedy & Specific & Path planning & LIF & Rate & Graph & STDP & MATLAB \\
\hline 
\cite{steffen2020networks} & Greedy & Specific & Path planning & LIF & Rate & Graph & STDP & NEST \\
\hline 
\cite{chao2023brain} & Greedy & Specific & Path planning & LIF & Rate & Graph & STDP & Python \\
\hline 
\cite{mniszewski2019graph} & SA & Specific & QUBO, GP & LIF & Rate & Recurrent & Noise & TrueNorth \\
\hline 
\cite{alom2017quadratic} &  LS & Specific & QUBO & LIF & Rate & Recurrent & Noise & TrueNorth \\
\hline
\cite{pierro2024solving} & SA & Specific & QUBO & LIF & Rate & Recurrent  & Noise & Loihi2 \\
\hline
\cite{henke2023sampling} & LS & Specific & QUBO & NEBM & Rate & WTA & Noise & Loihi2 \\
\hline 
\cite{fonseca2017using} & LS & Specific & CSP & LIF & Rate & WTA & Noise & Spinakker \\
\hline
\cite{sasaki2023swarm} & PSO & General & COP & IF OSNN & Time & Recurrent & Oscillators & Simulation \\
\hline 
\cite{fang2022solving} & Swarm & Specific & QUBO & LIF & Rate & Recurrent & Noise & Matlab \\
\hline
\cite{fang2019swarm} & ACO & Specific & TSP, COP & Many & Time & Recurrent & Pheromone & Ferroelectric \\
\hline
\cite{lele2023neuromorphic} & Swarm & Specific & QUBO & LIF & Rate & Recurrent & Noise & RRAM \\
\hline
\cite{liang2019neuromorphic} & LS & Specific & CSP & LIF & Rate & WTA  & Noise & DYNAP \\
\hline 
\cite{talbi2025NEVA} & EA & General & BOP & Individual & Rate & Population & Variation & LAVA \\
\hline 
\cite{cruz2025} & Meta & General & COP & Izhikevich & Time & Population & Oscillators & LAVA \\
\hline 
\end{tabular}
\caption{Some illutrative examples of Nheuristics.}
\label{tab:classification}
\end{table}

\section {Conclusions and perspectives}
\label{sec:conclusion}

NC technology is not yet ready for large-scale industrial deployment \cite{kudithipudi2025neuromorphic}, but it is expected to become a major area of focus in the coming decade. This presents a unique opportunity for researchers, as the successful application of NC will depend on key breakthroughs in specific areas. Advancing this technology requires sustained innovation, and we anticipate that future discoveries will drive its transition from research to real-world impact. This paper explores the potential of NC beyond cognitive applications, specifically in solving optimization problems. Neuromorphic-based metaheuristics (Nheuristics) offer a promising alternative for tackling complex optimization tasks while operating under constraints such as latency, power consumption, size, and weight—making them particularly suitable for embedded systems and edge devices in the IoT.

\medskip

The development of Nheuristics should be strategically positioned along a spectrum that balances biological plausibility with electronic feasibility. When designing Nheuristics, several key perspectives can be considered:
\begin{itemize}
\item Designing {\it hybrid Nheuristics} which combine population-based Nheuristics, such as Evolutionary Algorithms (EA) and Particle Swarm Optimization (PSO), with stochastic local search presents a promising strategy for enhancing the efficiency of Nheuristics. An initial attempt in this direction was introduced through memetic Nheuristics, as proposed in \cite{talbi2025NEVA}.

\item Most existing Nheuristics have been designed for small to medium-scale optimization problems, while tackling large-scale optimization remains a significant challenge. Research suggests that the brain exhibits structured, hierarchical connectivity, often described through fractal organization. This presents an exciting opportunity to design large-scale Nheuristics using decomposition-based strategies. A hierarchical architecture of SNNs could be leveraged to solve sub-problems efficiently, with concepts from fractal optimization and co-evolutionary algorithms potentially integrated within the SNN framework.
 
\item Addressing various classes of optimization problems, including {\it mixed optimization}, {\it multi-objective optimization}, and {\it optimization under uncertainty}, remains a crucial challenge. Certain metaheuristic approaches can be readily adapted to these problem types \cite{talbi2024metaheuristics}\cite{talbi2012multi}\cite{juan2023review}. When dealing with optimization problems involving {\it expensive objective function} (e.g., black-box), SNNs present a compelling solution by efficiently approximating complex nonlinear functions \cite{zaman2015function}.
\end{itemize} 

\medskip

Regarding the implementation of Nheuristics, several promising directions warrant further exploration:
\begin{itemize}
\item {\it Hardware-aware optimization} allows to focus on designing and adapting Nheuristics to maximize efficiency based on the constraints and capabilities of specific hardware platforms. This approach ensures optimal performance in terms of speed, power consumption, memory usage, and computational efficiency, particularly for deployment on resource-constrained devices such as edge computing systems, embedded processors, FPGAs, and neuromorphic chips.
 
\item The future of high-performance computing involves leveraging {\it extremely heterogeneous architectures} that integrate neuromorphic, quantum, and digital hardware (e.g., CPU, GPU) to solve complex optimization problems efficiently \cite{vetter2017architectures}. Each of these computing paradigms has unique advantages, and their synergistic integration can provide unprecedented computational capabilities.
\end{itemize} 

\bibliography{reference}
\bibliographystyle{abbrv}
\end{document}